\documentclass[review,number,12pt]{elsarticle}%

\usepackage{natbib}
\usepackage{geometry}
\usepackage{graphicx}
\usepackage{subfigure}
\usepackage{scrtime}
\usepackage{epstopdf}
\usepackage{enumerate}
\usepackage{amsmath}
\usepackage{siunitx}
\usepackage{amssymb}
\usepackage{upgreek}
\usepackage{appendix}
\usepackage[colorlinks, linkcolor=black, citecolor=black, anchorcolor=black]{hyperref}
\hypersetup{hidelinks}
\usepackage{algorithm}
\usepackage{algorithmicx}
\usepackage{algpseudocode}

\usepackage{multirow}
\usepackage{lineno}
\usepackage{enumerate}
\usepackage[section]{placeins}
\usepackage{pifont}

\usepackage{booktabs}
\usepackage{threeparttable}

\usepackage{booktabs}
\usepackage[dvipsnames]{xcolor}
\usepackage{tcolorbox}
\usepackage{soul}
\usepackage{makecell}
\usepackage{mathtools}
\usepackage{csquotes}

\def\romannumer#1{\uppercase\expandafter{\romannumeral#1}}

\makeatletter
\def\ps@pprintTitle{
	\let\@oddhead\@empty
	\let\@evenhead\@empty
	\def\@oddfoot{\centerline{\thepage}}%
	\let\@evenfoot\@oddfoot}
\makeatother

\begin{document}

	\begin{frontmatter}
		\title{Self-adaptive weights based on balanced residual decay rate for physics-informed neural networks and deep operator networks}
		
		\author[1]{Wenqian Chen}
		\ead{wenqian.chen@pnnl.gov}
		\author[1]{Amanda A. Howard}
		\ead{amanda.howard@pnnl.gov}
		\author[1]{Panos Stinis \corref{cor1}}
		\ead{panos.stinis@pnnl.gov}
		
		
		\cortext[cor1]{Corresponding author}
		
		\address[1]{Advanced Computing, Mathematics and Data Division \\
			Pacific Northwest National Laboratory \\ Richland, WA 99354, USA}

		%
		
		\begin{abstract}
			Physics-informed deep learning has emerged as a promising alternative for solving partial differential equations. However, for complex problems, training these networks can still be challenging, often resulting in unsatisfactory accuracy and efficiency. In this work, we demonstrate that the failure of plain physics-informed neural networks arises from the significant discrepancy in the convergence rate of residuals at different training points, where the slowest convergence rate dominates the overall solution convergence. Based on these observations, we propose a pointwise adaptive weighting method that balances the residual decay rate across different training points. 
			The performance of our proposed adaptive weighting method is compared with current state-of-the-art adaptive weighting methods on benchmark problems for both physics-informed neural networks and physics-informed deep operator networks. Through extensive numerical results we demonstrate that our proposed approach of balanced residual decay rates offers several advantages, including bounded weights, high prediction accuracy, fast convergence rate, low training uncertainty, low computational cost, and ease of hyperparameter tuning.
		\end{abstract}

		
	\end{frontmatter}
	\section{Introduction}
	Benefiting from the rapid advancements in computational capabilities, optimization algorithms, and automatic differentiation technologies \cite{paszke2019pytorch, abadi2016tensorflow}, physics-informed neural networks (PINNs)\cite{raissi2019physics} have emerged as a powerful tool for addressing both forward and inverse problems associated with partial differential equations (PDEs). Integrating physical laws directly into their framework, PINNs optimize a loss function that includes data and equation residuals to assist models in adhering to the underlying physical principles. Building upon the foundational concepts of PINNs, and the deep operator network (DeepONet) architecture \cite{lu2021learning}, physics-informed deep operator networks (PIDeepONets) extend these methodologies to the solution operators of  PDEs\cite{wang2021learning, goswami2022physics,goswami2023physics}. Both PINNs and PIDeepONets have enjoyed success in various settings, but they can still face convergence/accuracy issues \cite{wang2021understanding,wight2020solving,mcclenny2020self}. To mitigate these issues, various enhancements to the plain PINN/PIDeepONet or  approach have been proposed e.g., improved  network model \cite{wang2021understanding, wang2022improved}, adaptive sampling \cite{gao2023failure, tang2023pinns, wu2023comprehensive}, domain decomposition \cite{heinlein2021combining,heinlein2024multifidelity}, multi-fidelity learning \cite{chen2024feature, howard2023multifidelity, meng2020composite}, continual learning \cite{ howard2024multifidelity}, adaptive activation \cite{jagtap2020adaptive}, and Fourier feature embedding \cite{tancik2020fourier, wang2021eigenvector}.  In the current work, we focus on a self-adaptive weighting method designed to dynamically balance the training process of PINNs and PIDeepONets, aiming to improve their performance.
	
	Adaptive weighting in PINNs and PIDeepONets has revolutionized the way these models handle the training process by dynamically adjusting the weights assigned to different terms in the loss function. This method effectively balances the loss contributions from various parts of the domain, thereby addressing one of the fundamental challenges: ensuring that all physical laws are learned equally well without bias towards simpler or more dominant features. As a result,  improvement in convergence rates and increase in the accuracy and stability of the solutions has been obtained. 
	
	There are numerous adaptive weighting approaches, with the strategies for tuning weights varying considerably among approaches.  For instance, Wang et al. \cite{wang2021understanding}
	introduced a learning rate annealing algorithm that updates weights inversely proportional to the back-propagated gradients. Wang et al. \cite{wang2024respecting}  also defined a causal training for time-dependent problems that assigns larger weight to the loss functions contributions in a time-ordered fashion. Mattey and Ghosh \cite{mattey2022novel} solve sequentially in temporal subdomains with a plain PINN, using weights to penalize the departure from the already obtained solutions from previous training. Another popular strategy is to update weights positively proportional to (normalized) residuals. For instance, Liu and Wang \cite{liu2021dual} proposed a mimimax method to update weights, using gradient descent for the network parameters and gradient ascent for the weights (the gradient is proportional to residuals). McClenny and Braga-Neto \cite{mcclenny2020self} proposed a general variant of the mimimax method by employing pointwise weights instead of component-wise weights. Taking this mimimax strategy further, Song et al. \cite{song2024loss} and Zhang et al. \cite{zhang2023dasa} employed  auxiliary networks to represent the pointwise weights.  Anagnostopoulos et al. \cite{anagnostopoulos2024residual} proposed to update weights according to normalized residuals. A Lagrange
	multiplier-based method has also been employed for designing adaptive weights,  where the weights, namely the Lagrange multipliers applied to the constraint terms, are updated based on the residuals of constraint terms. 
	Basir et al. \cite{basir2022physics} proposed using the augmented Lagrangian method (ALM) to constrain the solution of PDE with boundary conditions or any available data. They then introduced an adaptive ALM \cite{basir2023adaptive} to enhance this approach, and further improved its capability with adaptive pointwise multipliers and the design of a dual problem \cite{basir2023investigating}. Son et al. \cite{son2023enhanced} proposed an augmented Lagrangian relaxation method for PINN training using pointwise multipliers. Neural tangent kernel (NTK)-based weighting is another strategy which updates the weights inversely proportionally to the eigenvalues of the NTK matrix. Wang et al. \cite{wang2022and} first proposed the NTK weighting method for PINN training,  and then extended it to PIDeepONet training \cite{wang2022improved}. The conjugate kernel (CK) has recently emerged as a faster alternative to NTK weights for PIDeepONet training, with similar accuracy \cite{howard2024conjugate}.

	In this work, we use a novel strategy to design self-adaptive weights. We begin with a toy problem to identify the failure mechanisms of plain physics-informed neural networks. Our testing uncovers two key observations: first, the convergence rates of residuals at various points differ significantly, spanning several orders of magnitude; second, the slowest convergence rate among all residuals predominantly dictates the overall solution convergence to the true values. Building on these insights, we propose a self-adaptive weighting method aimed at balancing the convergence rates of residuals by assigning greater weights to those with slower convergence rates. We also enforce that the average of all weights is 1, ensuring that the adaptive weights are bounded. Our numerical experiments with both PINNs and PIDeepONets indicate that our self-adaptive weighting method achieves high prediction accuracy, high training efficiency, and low training uncertainty.
	
	The rest of the paper is structured as follows. Section \ref{sec_failure_PINN} introduces the notion of \enquote{inverse residual decay rate} to describe the convergence rate of residuals and  uncover the failure mechanism of plain PINNs. Based on the observations in Section \ref{sec_failure_PINN}, a self adaptive weighing method based on the   balanced residual decay ratio (BRDR) for physics-informed machine learning is proposed and extended to mini-batch training in Section \ref{sec_SA}. The proposed self-adaptive weighting method is tested on physics-informed neural networks and physics-informed deep operator networks in Sections \ref{sec_test_PINN} and \ref{sec_test_DeepOnet}, respectively. To promote reproducibility and further research, the code and all accompanying data are available on {\bf{github.com/pnnl/ET-PINN.}}

	\section{Understanding the plain PINN failure mechanism}\label{sec_failure_PINN}
	\subsection{Physics-informed neural networks}
	Physics-informed neural networks (PINNs) aim at inferring a function $\mathbf{u}(\mathbf{x})$ of a system with (partially) known physics, typically defined in the form of partial differential equations (PDEs):
	\begin{equation}
		\begin{aligned}
			\mathcal{R}(\mathbf{u}(\mathbf{x})) =0&, \qquad \mathbf{x}\in\Omega\\
			\mathcal{B}(\mathbf{x}) =0&, \qquad \mathbf{x}\in \partial \Omega \\
		\end{aligned}
	\end{equation} 
	where $\mathbf{x}\in \mathbb{R}^{n_p}$ are $n_p$-dimensional spatial/temporal coordinates, $\mathcal{R}$ is a general partial differential operator defined on the domain $ \Omega$ and $\mathcal{B}$ is a general boundary condition operator defined on the boundary $ \partial \Omega$. For time-dependent problems, time $t$ is considered as a  component of $\mathbf{x}$, $\Omega$ is a space-time domain, and the initial condition will be assumed as a special boundary condition of the space-time domain. In PINNs, the solution $u(\mathbf{x})$ is first approximated as $u_{NN}(\mathbf{x}; \pmb{\theta})$ by a neural network model built with a set of parameters $\pmb{\theta}$.
	The partial derivatives of $u_{NN}(\mathbf{x}; \pmb{\theta})$ required for the estimation of the action of the operators $\mathcal{N}$ and $\mathcal{B}$ are readily computed by automatic differentiation. The training of the PINN is a multi-objective optimization problem aiming to minimize the residuals of the PDE and the boundary conditions, which are usually evaluated on a set of collocation points. In the plain PINNs, the optimizing objective, namely the loss function,  is defined as a linear combination of the square of the following residuals: 
	\begin{equation}\label{eq_loss_PINN}
		\begin{aligned}
			\mathcal{L}(\pmb{\theta}) 
			&=\frac{1}{N_{R}}\sum_{i=1}^{N_{R}}{\mathcal{R}^2(\mathbf{x}_{R}^i)} 					 +\frac{1}{N_{B}}\sum_{i=1}^{N_{B}}{\mathcal{B}^2(\mathbf{x}_{B}^i)}	  
		\end{aligned}
	\end{equation}
	where $\mathbf{x}_{R}=\{\mathbf{x}_{R}^i\}_{i=1}^{N_{R}} \subset \Omega$ are collocation points within the domain,  and $\mathbf{x}_{B}=\{\mathbf{x}_{B}^i\}_{i=1}^{N_{B}} \subset \partial\Omega$ are boundary points. 
	
	Usually, the loss function in Eq. \eqref{eq_loss_PINN} is minimized by gradient-based optimization algorithms, such as Adam \cite{kingma2014adam}. Ideally,  in the case of infinite residual/boundary points, if the loss function drops down to zero, all the residuals drop to zeros too and thus the system is solved exactly. However, limited by the number of residual/boundary points, the network approximating capability and the optimization error, the loss function cannot drop down to zero, and we can only try to minimize it as close to zero as possible.
	
	\subsection{Training dynamic of unweighted PINNs  } \label{sec_unWeightedPINN}
	Let us first consider an unweighted loss function
	\begin{equation}\label{eq_loss_PINN_unweighted}
		\begin{aligned}
			\mathcal{L}(\pmb{\theta}) 
			&=\sum_{i=1}^{N_{R}}{\mathcal{R}^2(\mathbf{x}_{R}^i; \pmb{\theta})} 					 +\sum_{i=1}^{N_{B}}{\mathcal{B}^2(\mathbf{x}_{B}^i; \pmb{\theta})}   
		\end{aligned}
	\end{equation}
	The training process of physics-informed neural networks can be described using the Neural Tangent Kernel (NTK) theory \cite{wang2022and}:
	\begin{equation}\label{eq_NTK}
		\left[ \begin{array}{c} \frac{d\mathcal{R}(\mathbf{x_R}; \pmb{\theta}(t))}{dt}\\ \frac{d\mathcal{B}(\mathbf{x_B}; \pmb{\theta}(t))}{dt}\end{array}  \right]
		= - 2 K(t)
		\left[ \begin{array}{c} \mathcal{R}(\mathbf{x_R}; \pmb{\theta}(t))\\ \mathcal{B}(\mathbf{x_B}; \pmb{\theta}(t))\end{array} \right]    	
	\end{equation}
	where $K(t)$ is the NTK matrix at training time $t$  defined as
	\begin{equation}
		K(t) = \left[\begin{array}{cc}
			K_{RR}(t) \quad K_{RB}(t) \\
			K_{BR}(t) \quad K_{BB}(t) 
		\end{array} \right]	 
	\end{equation}
	The entries of the NTK matrix are defined as follows:
	\begin{equation}
		\begin{aligned}
			\left[K_{RR}(t)\right]_{ij} &= \frac{d\mathcal{R}(\mathbf{x}_R^i; \pmb{\theta}(t))}{d\pmb{\theta}}\cdot \frac{d\mathcal{R}(\mathbf{x}_R^j; \pmb{\theta}(t))}{d\pmb{\theta}}, \qquad &1\le i,j \le N_R \\
			\left[K_{RB}(t)\right]_{ij} &= \frac{d\mathcal{R}(\mathbf{x}_R^i; \pmb{\theta}(t))}{d\pmb{\theta}}\cdot \frac{d\mathcal{B}(\mathbf{x}_B^j; \pmb{\theta}(t))}{d\pmb{\theta}}, \qquad &1\le i \le N_R, 1\le j \le N_B,  \\	
			\left[K_{BB}(t)\right]_{ij} &= \frac{d\mathcal{B}(\mathbf{x}_B^i; \pmb{\theta}(t))}{d\pmb{\theta}}\cdot \frac{d\mathcal{B}(\mathbf{x}_B^j; \pmb{\theta}(t))}{d\pmb{\theta}}, \qquad &1\le i,j \le N_B 
		\end{aligned}
	\end{equation}

	As demonstrated in \cite{jacot2018neural,wang2022and},  when the network width tends to infinity and  the learning rate tends to zero, the NTK matrix converges to a deterministic constant kernel, namely $K(t) \to K^*$. Substituting the eigendecomposition  $K^* =  Q \Lambda Q^T$ into Eq. \eqref{eq_NTK}, we have 
	\begin{equation}\label{eq_QR}
		Q^T\left[ \begin{array}{c} \mathcal{R}(\mathbf{x_R}; \pmb{\theta}(t))\\ \mathcal{B}(\mathbf{x_B}; \pmb{\theta}(t))\end{array} \right]  \approx \exp(-2 \Lambda t) Q^T \left[ \begin{array}{c} \mathcal{R}(\mathbf{x_R}; \pmb{\theta}(0))\\ \mathcal{B}(\mathbf{x_B}; \pmb{\theta}(0))\end{array} \right] 
	\end{equation}
	
	\begin{equation}\label{eq_QR2}
		\left[ \begin{array}{c} \mathcal{R}(\mathbf{x_R}; \pmb{\theta}(t))\\ \mathcal{B}(\mathbf{x_B}; \pmb{\theta}(t))\end{array} \right]  \approx Q\exp(-2 \Lambda t) Q^T \left[ \begin{array}{c} \mathcal{R}(\mathbf{x_R}; \pmb{\theta}(0))\\ \mathcal{B}(\mathbf{x_B}; \pmb{\theta}(0))\end{array} \right] 
	\end{equation}
	
\begin{figure}[!ht]
		\centering
		\includegraphics[ width=15cm, trim=2.0cm 0cm 3cm 0cm, clip=true]{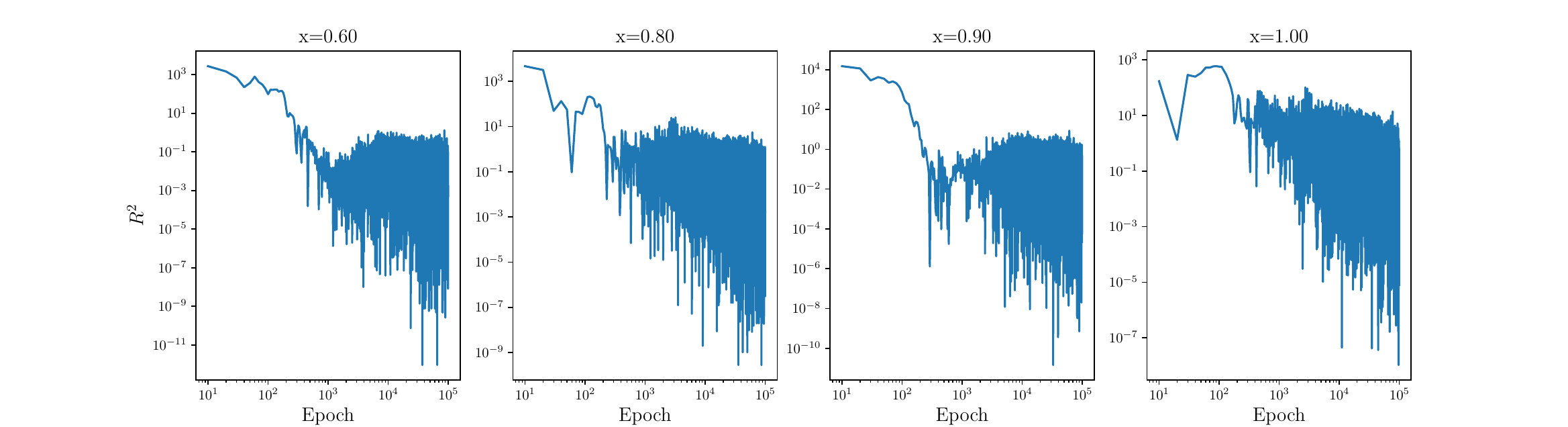}
		\caption{The residual decay process for four training points in a 1-dimensional Poisson equation, as described in Section \ref{sec_rdr_diff}.}
		\label{fig_residualDecay}
	\end{figure}

	Since $K^*$ is a positive semi-definite matrix, the eigenvalues of $K^*$ are all non-negative real numbers. This means that the $i$th entry of $Q^T\left[  \mathcal{R}(\mathbf{x_R}; \pmb{\theta}(t)), \mathcal{B}(\mathbf{x_B}; \pmb{\theta}(t))\right]^T $ decays approximately exponentially at a constant convergence rate $2\Lambda_i$. Therefore, the  evolution of the $i$th residual  in $\left[  \mathcal{R}(\mathbf{x_R}; \pmb{\theta}(t)), \mathcal{B}(\mathbf{x_B}; \pmb{\theta}(t))\right]^T $ is a linear combination of those decaying exponentials.
	It is reasonable to assume that the $i$th residual also decays exponentially in a short period, and its convergence rate falls between the $2\min(\Lambda)$ and $2\max(\Lambda)$. Note that this conclusion also applies to PIDeepONet. For details on the NTK theory of PIDeepONet, we refer the reader to \cite{wang2022improved}.  Figure \ref{fig_residualDecay} provides an example illustrating the residual decay process during training.

	It is not trivial to calculate the convergence rate of the residual at a training point, since it is always iteration-dependent. To describe the dynamic of residuals, we introduce a notion called \enquote{inverse residual decay rate}, which is used to indicate the convergence/decay rate of a residual. The inverse residual decay rate $irdr$ is defined as the ratio of the residual's square to its exponential moving average, namely
	\begin{equation}\label{eq_irdr}
		irdr= R^2/\sqrt{\overline{R^4}+eps}
	\end{equation} 
	where $eps$ is a tiny positive real number used to avoid division by zero, and it is opted as $eps=1E-14$. The exponential moving average $\overline{R^4}$ at training iteration $n$ is updated according to the following equation:
	\begin{equation}
		\overline{R_n^4} = \beta_c \overline{R_{n-1}^4} + (1-\beta_c) R_n^4
	\end{equation}
	where $\beta_c$ is the smoothing factor. A larger value of $\beta_c$ corresponds to a longer-period average, thereby placing more weight on past observations and less on the most recent observation.

	We note that the employment of the fourth-order residual, namely the 4th moment,  in Eq. \eqref{eq_irdr}—inside the squared root—is intended to make the instantaneous $irdr$ more responsive to large residuals. Intuitively, higher-order moments (such as the 4th moment) place greater emphasis on larger deviations from the mean, thereby highlighting scenarios where residuals are particularly problematic or persistent. Although we experimented with the 2nd moment as well, we found no decisive empirical advantage of one over the other. Both the 2nd and 4th moments can serve as approximations of the $irdr$. Given these considerations, we opted to use the 4th moment to enhance sensitivity to large residuals, while acknowledging that this choice may not universally yield a significant improvement over the 2nd moment.

	It is worth noting that the residual does not always strictly decrease during the training process, as shown in Fig. \ref{fig_residualDecay}. However, changes in the residual are reflected in the magnitude of $irdr$. In particular, if the residual decreases, we have $irdr < 1$; if the residual increases, then $irdr > 1$; and if the residual remains roughly unchanged, $irdr \approx 1$. Consequently, $irdr$ serves as a useful indicator of how the residual evolves and its convergence behavior.

	To intuitively visualize the relationship between the inverse residual decay rate $irdr$ and the convergence rate $\lambda$, we assume that the residual at a given training point decays exponentially with respect to the iteration index $n$ as $R = R_0\exp(-\lambda n)$, where $\lambda > 0$ is the convergence rate. The relationship between the inverse residual decay rate $irdr$ and the convergence rate $\lambda$ can be calculated numerically.
	
\begin{figure}[!ht]
		\centering
		\includegraphics[ width=12cm, trim=0cm 0cm 0cm 0cm, clip=true]{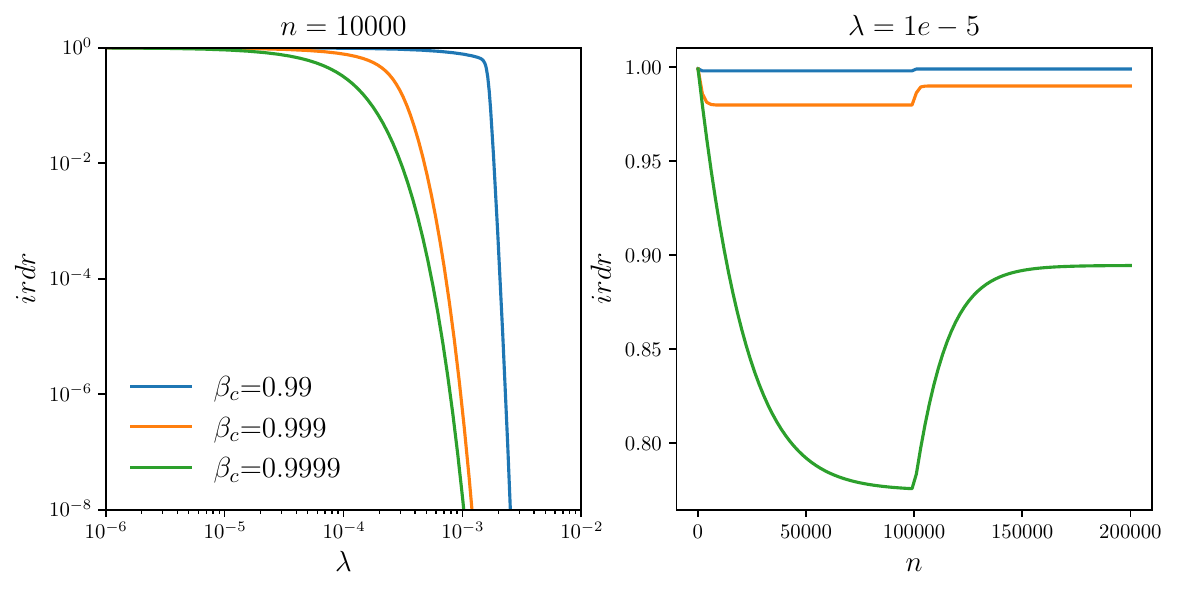}
		\caption{The relationship between the convergence rate $\lambda$ and the inverse residual decay rate ($irdr$) calculated with different smoothing factor $\beta_c$. The left panel shows $irdr$ for a time-decaying residual with a fixed convergence rate, $R = R_0 \exp(-\lambda n)$. The right panel illustrates $irdr$ for a time-decaying residual where the convergence rate is initially $\lambda = 1e-5$ and is reduced by half at $n = 100,000$.}
		\label{fig_rdr_lamda}
	\end{figure}

	The relationship between $\lambda$ and $irdr$ at $n = 10000$ is depicted in Fig. \ref{fig_rdr_lamda}(left). It is observed that $irdr$ is negatively proportional to $\lambda$ when $\lambda$ is large. Conversely, $irdr$ quickly increases to 1 when $\lambda$ is small. Additionally, for larger values of $\beta_c$, $irdr$ increases to 1 at smaller $\lambda$. Hence, a larger $\beta_c$ can be utilized to sense a broader range of $\lambda$. However, it is not necessarily true that a larger $\lambda$ is preferable. Consider another decay process for the residual:
	\begin{equation}
		R=\left\{
		\begin{aligned}
			&R_0\exp(-\lambda n) &\qquad &n\le 100000 \\
			&R_0\exp(- 100000\lambda)\exp(-0.5\lambda (n-100000)) &&n> 100000 \\
		\end{aligned}
		\right.
	\end{equation}
	where $\lambda = 1 \times 10^{-5}$. In this scenario, the convergence rate is higher during the initial stage and then transitions to a lower convergence rate. This pattern mimics the actual training process, where the convergence rate typically decreases with an increasing number of epochs. The calculated $irdr$ for this process is shown in Fig. \ref{fig_rdr_lamda}(right). When $\beta_c$ is small, $irdr$ can rapidly respond to changes in convergence. 
	
	From these observations, we conclude that $\beta_c$ should neither be too small nor too large so that we can use $irdr$ to accurately identify the convergence rate and adapt to its variations.

	\subsection{Vast convergence disparities can lead to failure of plain PINNs} \label{sec_rdr_diff}
	Plain PINNs have been observed to have convergence issues for problems with sharp space/time transitions \cite{chen2024feature}. As an example, consider the 1D Poisson equation:
	\begin{equation}
		\begin{aligned}
			&\frac{\partial^2 u}{\partial x^2}=f(x), \qquad x\in[0,1] \\
			&u(0)=u(1)=0
		\end{aligned}
	\end{equation}
	with the artificial solution $u(x)=\sin(2 k \pi  x^2)$. 
	The oscillating frequency of the solution is 0 at $x=0$, and  increases to $k$ at $x=1$. The frequency discrepancy is more apparent with increasing $k$. With increasing $k$, resolving the high-frequency oscillation as well as the vast frequency discrepancy present  challenges for the plain PINN. 
	
	To approximate the solution, we use $u(x; \pmb{\theta})$ which is a 6-layer fully connected neural network with 50 neurons per layer and the hyperbolic tangent activation function. The loss function is defined as 
	\begin{equation}
		\mathcal{L}(\pmb{\theta}) = \frac{1}{2}\left( u^2(0; \pmb{\theta})+u^2(1; \pmb{\theta}) \right)
		+\frac{1}{N_R}\sum_{i=1}^{N_R}\left(\frac{\partial^2 u(x_i; \pmb{\theta})}{\partial x^2}-f(x_i)\right)^2.
	\end{equation}
	where $N_R=1000$ uniform residual points are sampled in the domain $[0,1]$. The loss function is minimized by Adam optimizer with 100000 full-batch training steps using a constant learning rate 0.001. 
	The  prediction errors of the plain PINN for $k=2,4,8$ are given in Table \ref{table_err_1DPoisson}. The reported prediction error is  the relative $L_2$ error defined as follows
	\begin{equation}
		\epsilon_{L_2} = \frac{\|u-u_E\|_2}{\|u_E\|_2}
	\end{equation}
	where $u$ and $u_E$ are vectors of predicted solutions and the exact solutions evaluated on 10000 uniform sampled points, respectively. It is shown that the performance of the plain PINN degrades with the increase of $k$.

	\begin{table}[tbp]
		\newcommand{\tabincell}[2]{\begin{tabular}{@{}#1@{}}#2\end{tabular}}	
		\centering
		\caption{Prediction errors of the plain PINN and the balanced-residual-decay-rate (BRDR) PINN for the Poisson equation. Note that the BRDR method will be detailed in Section \ref{sec_SA}}.
		\begin{tabular}{cccc}
			\toprule
			Method & $k=2$ &  $k=4$ &  $k=8$ \\
			\midrule		
			Plain & $(9.70\pm 4.61)\times10^{-3}$ &   $(9.21\pm 9.31)\times10^{-2}$ & $(1.27\pm 1.03)\times10^{0}$  \\
			BRDR & $(7.91\pm 4.70)\times10^{-4}$ & $(2.59\pm 1.24)\times10^{-3}$ & $(1.07\pm 0.70)\times10^{-2}$ \\
			\bottomrule
		\end{tabular}
		\label{table_err_1DPoisson}
	\end{table}

\begin{figure}[!ht]
		\centering	
		\subfigure[Solution $u$]{
			\label{Fig_poisson_sol}
			\includegraphics[width=15cm, trim=0cm 0.2cm 0cm 0.2cm, clip=true]{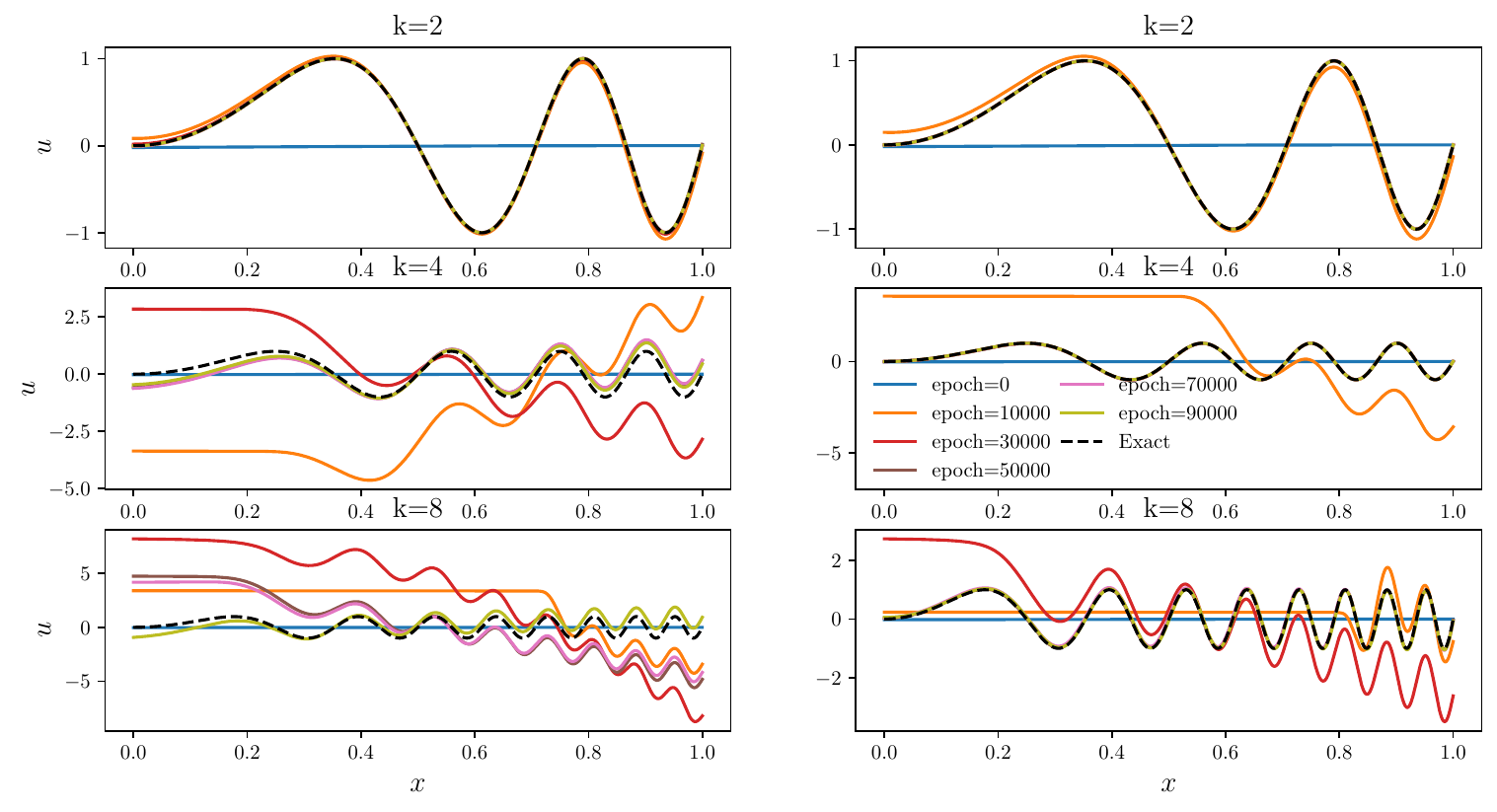}
			\label{fig_Poisson_solution}
		}	
		
		\subfigure[Average inverse residual decay rate  $\overline{irdr}$]{
			\label{Fig_poisson_rdr}
			\includegraphics[width=15cm, trim=0cm 0.2cm 0cm 0.2cm, clip=true]{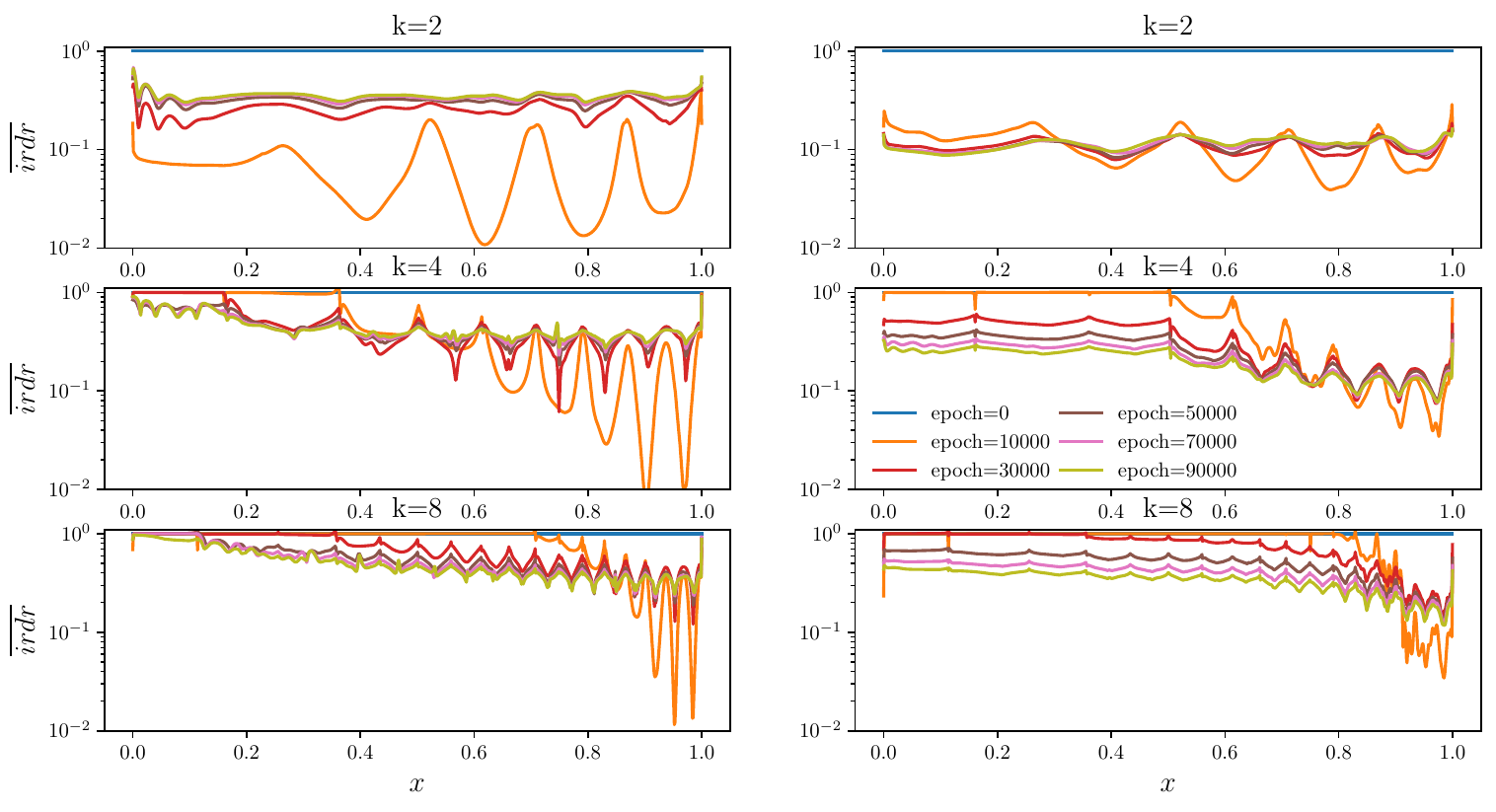}
			\label{fig_Poisson_rdr}
		}
		\caption{The history of solution and average inverse residual decay rate during training process from plain PINN (left) and BRDR PINN (right) for the 1D Poisson equation. Note that both the plain PINN and BRDR PINN share the same network initialization for the same $k$. }
	\end{figure}

	To uncover the reason behind the failure, let us take a look at the inverse residual decay rate $irdr$ at all the training points. For iteration $n$, we calculate the average of $irdr$ from the first iteration to the current iteration for each training point. The average inverse residual decay rate $\overline{irdr}$ for the plain PINN is illustrated in Fig. \ref{Fig_poisson_rdr}(left). It is shown that $\overline{irdr}$ can fluctuate by about two orders of magnitude over all the training points. Within the domain, the training points which correspond to larger values of $x$ have smaller average inverse residual decay rates, implying that the network tries to capture high-frequency oscillation first.\footnote{ We observe that this behavior represents an exception to the common spectral bias. This deviation is driven by the properties of the source term $f(x) = -16k^2\pi^2x^2\sin(2k\pi x^2) + 4k\pi\cos(2k\pi x^2)$, characterized by oscillations that not only increase in frequency but also in amplitude from left to right across the domain. These unique characteristics necessitate a shift in the learning focus of the network towards more extensively addressing these higher frequency components. } This is demonstrated by the evolution history of the predicted solution in Fig. \ref{Fig_poisson_sol}(left), where the left part of the solution is nearly flat at the early training stage, specifically for $k=4$ and $k=8$.
	Meanwhile, $\overline{irdr}$  at $x=0$ is close to 1 for $k=4$ and $k=8$, implying that the residual has almost no change during the training process, and thus the plain PINN fails to converge. For $k=4$ and $epoch=90000$, $\overline{irdr}$ is close to 1 but a little smaller than 1, so we can still observe in Fig. \ref{Fig_poisson_sol}(left) that the solution converges to the exact solution although at a very low rate. For $k=8$ and $epoch=90000$, $\overline{irdr}$ is much closer to 1, thus it converges to the exact solution at a much lower rate. Based on the above observations, we have the following conclusions:
	\begin{enumerate}[1]
		\item During the network training process, the residuals at different points may exhibit varying decay rates, with fluctuations in these rates spanning several orders of magnitude.
		\item The convergence rate of the predicted solution to the exact solution is primarily determined by the  largest inverse residual decay rate. A higher maximum inverse residual decay rate results in a slower convergence rate.
	\end{enumerate}
	
	In the next section, we introduce an adaptive weighting method aimed at balancing the residual decay rate (BRDR). The results from implementing the BRDR PINN are depicted in Fig. \ref{Fig_poisson_rdr}(right). The results indicate that the distribution of inverse residual decay rates becomes significantly more uniform, and the maximum inverse residual decay rate observed in the plain PINN is markedly reduced with the adaptive weighting method. As a consequence, the prediction converges to the exact solution more rapidly, and the final achieved error and uncertainty are considerably lower, as listed in Table \ref{table_err_1DPoisson}.

	\section{Physics-informed machine learning with balanced residual decay rate } 
	\label{sec_SA}
	\subsection{Physical insights of weighed PINNs}
	To build a general weighted PINN framework, we use a scaling factor and a set of normalized weights  with each weight assigned to a residual term, namely
	\begin{align}
		\mathcal{L}(\pmb{\theta}; \mathbf{w},s) &= s \left( \dfrac{1}{N_R}\sum_{i=1}^{N_R}w_R^i\mathcal{R}^2(\mathbf{x}_R^i)
		+\dfrac{1}{N_B}\sum_{i=1}^{N_B}w_B^i\mathcal{B}^2(\mathbf{x}_B^i)
		\right)	\label{eq_loss_weighted}\\
		&s.t. \qquad  \text{mean}({w}):=\frac{\sum_{i=1}^{N_R}w_R^i+\sum_{i=1}^{N_B}w_B^i}{N_R+N_B}=1 \label{eq_weight_constraint}
	\end{align}
	where $w_R^i>0$ is the weight assigned to each residual term, $w_B^i>0$ is the weight assigned to each boundary point, and $\mathbf{w}$ is the collection of these weights. 
	The scaling factor $s$ is employed to scale all the weights, so that the formulation could cover all kinds of possible weight distribution.

	The training process of  physics-informed neural networks with weights can be approximated as follows:
	\begin{equation}\label{eq_NTK_weighted}
		\left[ \begin{array}{c} \frac{d\mathcal{R}(\mathbf{x_R}; \pmb{\theta}(t))}{dt}\\ \frac{d\mathcal{B}(\mathbf{x_B}; \pmb{\theta}(t))}{dt}\end{array}  \right]
		= - 2 s K(t) \text{diag}(\mathbf{w})\text{diag}(1/\mathbf{N})
		\left[ \begin{array}{c} \mathcal{R}(\mathbf{x_R}; \pmb{\theta}(t))\\ \mathcal{B}(\mathbf{x_B}; \pmb{\theta}(t))\end{array} \right]    	
	\end{equation}
	where $\mathbf{N}=[N_R,...,N_R, N_B,...,N_B]$ and the definition of $K(t)$ is the same as in Section \ref{sec_unWeightedPINN}. The terms $1/N_R$ and $1/N_B$ can be considered user-defined weight constants. The introduction of $\mathbf{w}$ and $s$ modifies the training dynamics. Empirical observations indicate that increasing the weight at a specific training point enhances its convergence rate. Regarding the scaling factor, $s$ scales the eigenvalues of the matrix $K(t) \text{diag}(\mathbf{w}) \text{diag}(1/\mathbf{N})$, thereby influencing the overall convergence velocity. In the following sections, we will elucidate the procedure for updating $\mathbf{w}$ and $s$ during the network training process.

	\subsection{Balanced residual decay rate (BRDR)}\label{sec_brdr}
	As demonstrated in Section \ref{sec_rdr_diff}, the largest inverse residual decay rate dominates the convergence rate of the training process. To address this issue, we assign a larger weight to the training term with a larger inverse residual decay rate, namely,
	\begin{equation}
		\mathbf{w} \propto \mathbf{irdr}
	\end{equation}
	where $\mathbf{irdr}$ is  the collections of inverse residual decay rate $irdr$ for all the training terms. At training iteration $n$, to meet the normalization constraint on $\mathbf{w}$ in Eq. \eqref{eq_weight_constraint}, we set 
	\begin{equation}
		\mathbf{w}_{n}^{ref} =  \frac{\mathbf{irdr}_{n}}{\text{mean}(\mathbf{irdr}_{n})}
	\end{equation}	
	To filter out noise during network training, we employ an exponential moving average method to update the weights, namely,
	\begin{equation}\label{eq_weight_update}
		\mathbf{w}_{n} = \beta_w \mathbf{w}_{n-1} + (1-\beta_w)\mathbf{w}_{n}^{ref}
	\end{equation}
	where $\beta_w$ is a smoothing factor. This idea of using a normalized quantity and an exponential moving average for updating weights is also employed in the recently proposed residual-based attention (RBA) method, \cite{anagnostopoulos2024residual,shukla2024comprehensive}, which helps ensure that the weights remain bounded and vary smoothly.

	\subsection{Adaptive scaling factor}\label{sec_max_lr}
	For the loss function defined in Eq. \eqref{eq_loss_weighted}, the loss without scaling $\mathcal{L}_{-s}=\mathcal{L}/s$ satisfies the following ordinary differential equation
	\begin{equation}\label{eq_eta0}
		\begin{aligned}
			\frac{d \mathcal{L}_{-s}}{d t} &= \nabla_{\pmb{\theta}} \mathcal{L}^T_{-s} \cdot \frac{d \pmb{\theta}}{d t} + \nabla_{\mathbf{w}}\mathcal{L}_{-s} \cdot \frac{d \mathbf{w}}{d t}\\
			&= -\nabla_{\pmb{\theta}} \mathcal{L}^T_{-s} \cdot \nabla_{\pmb{\theta}}\mathcal{L} + \left[\frac{1}{N_R}\mathcal{R}^2(\mathbf{x}_R),\frac{1}{N_B}\mathcal{B}^2(\mathbf{x}_B)\right]^T\cdot\frac{d \mathbf{w}}{d t}
		\end{aligned}
	\end{equation}
	For the first term of Eq. \eqref{eq_eta0}, we replace ${d \pmb{\theta}}/{d t}$ with the total-loss gradient$ \nabla_{\pmb{\theta}} \mathcal{L}$ in accordance with the gradient‐flow formulation. 
		According to the constraint $\text{mean}(\mathbf{w})=1$ in Eq. \eqref{eq_weight_constraint}, we have $\text{mean}({d \mathbf{w}}/{d t})=0$. Therefore, we assume the second term in Eq. \eqref{eq_eta0} as zero. 
		Note that this assumption primarily simplifies the formula, since in practice the second term can fluctuate and may not strictly vanish.

	\begin{equation}\label{eq_eta}
		\begin{aligned}
			\frac{d \mathcal{L}_{-s}}{d t}
			&\approx -\nabla_{\pmb{\theta}} \mathcal{L}^T_{-s} \cdot \nabla_{\pmb{\theta}}\mathcal{L} 
			&=-1/s \left\|\nabla_{\pmb{\theta}}\mathcal{L}\right\|_2^2 
			&=-\frac{\left\|\nabla_{\pmb{\theta}}\mathcal{L}\right\|_2^2}{\mathcal{L}}\mathcal{L}_{-s}
		\end{aligned}
	\end{equation}
	
	For the stable numerical simulation of the ODE $y_t=-\lambda y$ with the Euler forward method, the time step $\Delta t$ should satisfy $\Delta t \le 2/\lambda.$ Applying the stability constraint to  Eq. \eqref{eq_eta}, we find
	\begin{equation}\label{eq_etaCriteria}
		\eta =\Delta t \le \frac{2\mathcal{L}}{\left\|\nabla_{\pmb{\theta}}\mathcal{L}\right\|_2^2}
	\end{equation}
	where $\eta$ is the learning rate. Since $\mathcal{L}$ is proportional to the scaling factor $s$,  tuning $s$ during the network training process can make $\eta$ stay close to its maximum limit, thus accelerating the training process. Based on this idea, given the training status at iteration  $n-1$, namely the loss $\mathcal{L}_{n-1}$ and its gradient $\nabla_{\pmb{\theta}}\mathcal{L}_{n-1}$, we can derive the maximum scaling factor $s_{n-1}^{max}$ as follows: 
	\begin{equation}
	\eta 		=\frac{2\mathcal{L}_{n-1}^*}{\left\|\nabla_{\pmb{\theta}}\mathcal{L}_{n-1}^*\right\|_2^2} 
	= \frac{2\frac{s_{n-1}^{max}}{s_{n-1}}\mathcal{L}_{n-1}}{\frac{(s_{n-1}^{max})^2}{(s_{n-1})^2}\left\|\nabla_{\pmb{\theta}}\mathcal{L}_{n-1}\right\|_2^2}
		=\frac{s_{n-1}}{s_{n-1}^{max}}\frac{2\mathcal{L}_{n-1}}{\left\|\nabla_{\pmb{\theta}}\mathcal{L}_{n-1}\right\|_2^2} 
	\end{equation}
	where $\mathcal{L}_{n-1}^*$ denotes the loss obtained by tuning the scaling factor $s$ to precisely satisfy the equality in inequality \eqref{eq_etaCriteria} above.
	So, we have
	\begin{equation}\label{eq_st_max}
		s_{n-1}^{max} = \frac{s_{n-1}}{\eta}\frac{2\mathcal{L}_{n-1}}{\left\|\nabla_{\pmb{\theta}}\mathcal{L}_{n-1}\right\|_2^2} 
	\end{equation} 
	Then we use the derived $s_{n-1}^{max}$ to update the scaling factor $s$ based on the exponential moving average method, namely
	\begin{equation}\label{eq_s_update}
		s_{n} = \beta_s s_{n-1} + (1-\beta_s)s_{n-1}^{max}
	\end{equation}
	where $\beta_s$ is a smoothing factor. Since the scaling factor $ s$ functions similarly to the learning rate $ \eta $, we propose updating $s $ synchronously with $ \eta $. For example, when the learning rate $\eta $ decreases, the update velocity of the scaling factor $ s $ is also expected to decrease. In the following tests, unless otherwise specified, we set $\beta_s = 1 - \eta$.

	We note that updating the scaling factor via Eqs. \eqref{eq_st_max} and \eqref{eq_s_update} adds almost no extra cost—since $\mathcal{L}_{n-1}$ and its gradient $\nabla_{\theta}\mathcal{L}_{n-1}$ are already computed during back-propagation. Multiplying the loss by this factor makes each gradient step larger when the factor is above 1 and smaller when it’s below 1. This produces a similar effect to Adam’s per-parameter step-size adaptation, though by a different mechanism. We have not yet fully understood how these two adjustments interact, but our ablation study (see \ref{sec_ablation}) shows that the scaling factor generally reduces prediction error across most test cases—especially when paired with pointwise weights (Section \ref{sec_brdr}). We therefore recommend using the scaling factor as a complementary component alongside pointwise weights, rather than applying it in isolation, since standalone use can sometimes cause slight performance degradation (see \ref{sec_ablation}). Throughout this paper, the scaling factor is applied by default unless otherwise noted.

	\subsection{Mini-batch training}\label{sec_mini_batch}
	Mini-batch training is commonly employed in physics-informed machine learning for several reasons: it helps manage computational resources by fitting training within memory constraints and enhances computational efficiency. It facilitates the use of stochastic gradient descent and its variants \cite{robbins1951stochastic,ruder2016overview}, enabling more frequent model updates which can lead to faster convergence and better handling of complex loss landscapes inherent in physics.
	
	In this work, we restrict ourselves only to the scenario where all the training points are pre-selected before training and a subset of training points is randomly chosen at each training step.
	The proposed method in Sections \ref{sec_brdr} and  \ref{sec_max_lr} can be extended to mini-batch training straightforwardly, except for the calculation of the exponential moving average of quantities. Since a specific training point cannot be chosen at every training step, it is too expensive to update the weights for all the points at each iteration. It is efficient to only update the weights associated to the chosen points at each training step. We assume $\Delta n_i$ is the training iteration interval for the $i$th training point, which is the difference of current step and the last previous step that the training point was chosen. The weights are then updated as
	
	\begin{equation}\label{eq_weight_updateMini}
		\mathbf{w}_{n,i} = \beta_w^{\Delta n_i} \mathbf{w}_{n-\Delta n_i} + (1-\beta_w^{\Delta n_i})\mathbf{w}_{n,i}^{ref}
	\end{equation}
	
	Similarly, the exponential moving average $\overline{R_{n,i}^4}$ for the residual at the $i$th training point  is calculated as follows:
	\begin{equation}
		\overline{R_{n,i}^4} =\beta_c^{\Delta n_i}\overline{R_{n-\Delta n_i,i}^4} + (1-\beta_c^{\Delta n_i})R_{n,i}^4
	\end{equation} 

	For updating of the scaling factor in Eq. \eqref{eq_s_update}, it is also calculated with exponential moving average. However, the scaling factor can be updated in Eq. \eqref{eq_s_update} directly at each training step without any modifications, since it is accessible at each training step.
	
	When the number of batches is very large (e.g., 1000), it is advisable to use a larger smoothing factor, such as setting it to 0.9999 instead of 0.999. This ensures that the effective smoothing factors $\beta_c^{\Delta n_i}$ and $\beta_w^{\Delta n_i}$ remain sufficiently large, allowing past quantities to continue exerting a meaningful influence.

	\subsection{Summary}
	Consider a physics-informed neural network $\mathbf{u}_{NN}(\mathbf{x}; \pmb{\theta})$ with training parameters $\pmb{\theta}$, and a weighted loss function 
	\begin{equation}
		\mathcal{L}(\pmb{\theta}; \mathbf{w},s, \pmb{\alpha}) = s \left( \frac{\alpha_R}{N_R}\sum_{i=1}^{N_R}w_R^i\mathcal{R}^2(\mathbf{x}_R^i)
		+\frac{\alpha_B}{N_R}\sum_{i=1}^{N_B}w_B^i\mathcal{B}^2(\mathbf{x}_R^i)
		\right)
	\end{equation}
	where $\mathbf{w}$ represents pointwise adaptive weights allocated to collocation points $\{\mathbf{x}_R^i\}_{i=1}^{N_R}$ and $\{\mathbf{x}_B^i\}_{i=1}^{N_B},$ $s$ is an adaptive scaling factor, and $\pmb{\alpha}=\{\alpha_R, \alpha_B\}$ are user-defined weight constants to normalize the residuals. According to our tests in Section \ref{sec_test_PINN}, although simply setting $\pmb{\alpha}=1$ could be enough to achieve rather accurate results, setting a specific  $\pmb{\alpha}$ can significantly improve prediction accuracy. In the following, we refer to training with  $\pmb{\alpha}=1$ as BRDR training, and training with specifically defined  $\pmb{\alpha}$ as BRDR+ training. To avoid any confusion, we clarify that we use the term "BRDR" as the name of our weighting method to highlight our primary contribution: balancing the residual decay rate. By default, the BRDR method incorporates both the adaptive pointwise weights (see Section \ref{sec_brdr}) and the adaptive scaling factor (see Section \ref{sec_max_lr}), unless otherwise specified.
	
	In summary, after the specification of the user-defined hyperparameters which include the learning rate $\eta$, the smoothing factors $\beta_c$ and $\beta_w$, the batch sizes $N_{Rb}$ and $N_{Bb}$, and the weight constants $\alpha_R$ and  $\alpha_B$, the training process can proceed as detailed in Algorithm \ref{algorithm_adaptive}. Note that the weights and scaling factor are all initialized at 1, namely $\mathbf{w}=s=1$. Although Algorithm \ref{algorithm_adaptive} is specifically employed for problems with two loss components (PDE loss and BC loss), its extension to multiple components is straightforward. 
	\begin{algorithm}
		\caption{Self-adaptive weighting based on balanced residual decay rates.}
		\label{algorithm_adaptive}
		\begin{algorithmic}
			\Statex \Comment{Initialization}
			\State $\mathbf{w} \gets 1; \quad s \gets 1 $;\quad $\hat{\mathcal{R}} \gets 0$;\quad $\hat{\mathcal{B}} \gets 0$; \quad $\mathbf{n}_{last} \gets 0$
			\For{$n \gets$ 1 \textbf{to} $n_{max}$}
			\Statex \Comment{Sample batch indices}
			\State
			$\{i_k\}_{k=1}^{N_{Rb}} \subset \{i\}_{i=1}^{N_R}; \qquad 
			\{i_k\}_{k=1}^{N_{Bb}} \subset \{i\}_{i=1}^{N_B}$
			\Statex \Comment{Forward propagation}
			\State $\mathcal{R}_{i_k} \gets \mathcal{R}(\mathbf{x}_R^{i_k}; \pmb{\theta})$	 	
			;\qquad $\mathcal{B}_{i_k} \gets \mathcal{B}(\mathbf{x}_B^{i_k}; \pmb{\theta})$  
			\Statex \Comment{Calculate effective smoothing factors}  
			\State $\beta_{c,eff} \gets \beta_c^{n-n_{i_k,last}}$;\quad $\beta_{w,eff} \gets \beta_w^{n-n_{i_k, last}}$; \quad $n_{i_k,last} \gets n$
			\Statex \Comment{Calculate the inverse residual decay rate $irdr$, denoted as $c$ for simplicity}
			\State $\hat{\mathcal{R}}_{i_k} \gets \beta_{c,eff}\hat{\mathcal{R}}_{i_k}  + (1-\beta_{c,eff}) \mathcal{R}_{i_k}^4$ 
			;\qquad $\hat{\mathcal{B}}_{i_k} \gets \beta_{c,eff} \hat{\mathcal{B}}_{i_k}  + (1-\beta_{c,eff}) \mathcal{B}_{i_k}^4$ 
			\State $c_{R,{i_k}} \gets  \dfrac{\mathcal{R}_{i_k}^2}{ \sqrt{\hat{\mathcal{R}}_{i_k}/(1-\beta_c^{n})}+eps}$
			;\qquad $c_{B,{i_k}} \gets  \dfrac{\mathcal{B}_{i_k}^2}  {\sqrt{\hat{\mathcal{B}}_{i_k}/(1-\beta_c^{n})}+eps}$    
			\State $\overline{c} \gets \dfrac{\sum_{k=1}^{N_{Rb}}{c_{R,{i_k}}} + \sum_{k=1}^{N_{Bb}}{c_{B,{i_k}}}}{N_{Rb}+N_{Bb}} $
			\Statex \Comment{Update weights}
			\State $w_R^{{i_k}} \gets \beta_{w,eff} w_R^{i_k} + (1-\beta_{w,eff})\dfrac{c_{R,{i_k}}}{\overline{c}}$  
			;\qquad $w_B^{{i_k}} \gets \beta_{w,eff} w_B^{i_k} + (1-\beta_{w,eff})\dfrac{c_{B,{i_k}}}{\overline{c}}$ 
			\Statex \Comment{Assemble the loss function}
			\State 	$\mathcal{L}(\pmb{\theta}; \mathbf{w},s, \pmb{\alpha}) =s\left( \dfrac{\alpha_R}{N_R}\sum_{k=1}^{N_{Rb}}w_R^{i_k}\mathcal{R}^2(\mathbf{x}_R^{i_k})
			+\dfrac{\alpha_B}{N_B}\sum_{k=1}^{N_{Bb}}w_B^{i_k}\mathcal{B}^2(\mathbf{x}_R^{i_k}) \right)$
			\Statex \Comment{Backward propagation}
			\State $\nabla_{\pmb{\theta}}\mathcal{L} \gets$ \text{Backward propagation}
			\Statex \Comment{Update the scaling factor with the smoothing factor $\beta_s=1-\eta$}
			\State $s_0 \gets s$
			;\qquad $s  \gets (1-\eta)s + \dfrac{2s \mathcal{L}}{\left\|\nabla_{\pmb{\theta}}\mathcal{L}\right\|_2^2}$
			\Statex \Comment{Correct the gradients}
			\State $\nabla_{\pmb{\theta}}\mathcal{L} \gets \dfrac{s}{s_0} \nabla_{\pmb{\theta}}\mathcal{L} $
			\Statex \Comment{Update the parameters with gradient descent}
			\State $\pmb{\theta} \gets \pmb{\theta} - \eta \nabla_{\pmb{\theta}}\mathcal{L} $
			\EndFor
		\end{algorithmic}
	\end{algorithm}

	\section{Numerical results for physics-informed neural networks}\label{sec_test_PINN}
	To validate the performance of the BRDR weighting method in training PINNs, we tested it on three benchmark problems: the 2D Helmholtz equation,  the 1D Allen-Cahn equation,  and the 1D Burgers equation. For comparison, we also report the error from training with fixed weights, the soft-attention (SA) weighting method \cite{mcclenny2020self}, and the residual-based attention (RBA) method \cite{anagnostopoulos2024residual}. The reported error is  defined as the 
	$L_2$ relative error:
	\begin{equation}
		\epsilon_{L_2} = \frac{\|u - u_E\|_2}{\|u_E\|_2}
	\end{equation}
	where $u$ and $u_E$ are vectors of the predicted solutions and the reference solutions on the test set, respectively.
	
	In this section, we use the mFCN network architecture (see  \ref{sec_net}) with 6 hidden layers, each containing 128 neurons. The hyperbolic tangent function is employed as the activation function. The network parameters are initialized using the Kaiming Uniform initialization \cite{he2015delving}. Specifically, for a module of shape (out\_features, in\_features), the learnable weights and biases are initialized from $\mathcal{U}(-\sqrt{k}, \sqrt{k})$, where $k = 1/\text{in\_features}$.
	We use only the Adam optimizer \cite{kingma2014adam} for updating the training parameters. Although the L-BFGS optimizer \cite{liu1989limited} can fine-tune network parameters further, it is known for its significant drawbacks, including high computational cost and instability, particularly in large-scale problems. Therefore, we have chosen not to use the L-BFGS optimizer. 
	All training procedures described in this section are implemented using PyTorch \cite{paszke2019pytorch}. Training computations were performed on a GPU cluster, with each individual training run utilizing a single NVIDIA\textsuperscript{\textregistered} Tesla P100 GPU. All computations were conducted using 32-bit single-precision floating-point format.

	\subsection{2D Helmholtz equation}
	The 2D Helmholtz equation is defined as follows:
	\begin{equation}
		\begin{aligned}
			\frac{\partial^2 u}{\partial x^2}
			+\frac{\partial^2 u}{\partial y^2}
			+k^2 u - q(x,y) &= 0, \qquad (x,y)\in [-1,1]^2 \\
			u(x,\pm 1)&=0,        \qquad  x\in[-1,1] \\
			u(\pm 1, y)&=0, \qquad y\in[-1,1] \\
		\end{aligned}
	\end{equation}
	with the manufactured solution  $u_E(x,y)=\sin(a_1\pi x)\sin(a_2 \pi y)$, where $k=1$, $a_1$=1 and $a_2=4$ is considered. $q(x,y)$ is the source term defined by
	\begin{equation}
		q(x,y) = \left(k^2-(a_1\pi)^2-(a_2\pi)^2\right)\sin(a_1\pi x)\sin(a_2 \pi y).
	\end{equation}

	The loss function is defined as follows:
	\begin{equation}
		\mathcal{L}(\pmb{\theta}; \mathbf{w},s,\pmb{\alpha}) = s \left( \frac{\alpha_R}{N_R}\sum_{i=1}^{N_R}w_R^i\mathcal{R}^2(\mathbf{x}_R^i)
		+\frac{\alpha_B}{N_B}\sum_{i=1}^{N_B}w_B^i\mathcal{B}^2(\mathbf{x}_B^i)
		\right),
	\end{equation}
	where $\mathcal{R}$ and $\mathcal{B}$ represent the PDE operator and the boundary condition (BC) operator, respectively.
	
	The choice of location of the training points and the BRDR training setup are provided in Table \ref{tab_setup_PINN}. For fixed-weight training, the weights for both the boundary conditions (BC) and partial differential equations (PDE) are set to 1. For the soft-attention (SA) training setup, we follow the configuration given in reference \cite{mcclenny2020self} for Adam training. The pointwise self-adaptive weights for BC and PDE are all initialized using uniform sampling $\mathcal{U}(0,1)$, and the weights are updated with a fixed learning rate of 0.005. For the residual-based attention (RBA) training setup, we use the configuration provided in reference \cite{anagnostopoulos2024residual}. In this setup, the weights for BC are fixed at 100, and the pointwise self-adaptive weights for PDE are initialized at 0. These weights are then updated with a decay rate of 0.9999, an offset of 0,  and a learning rate of 0.001.
	
	\begin{table}[htbp]
		\centering
		\caption{The choice of location of the training points and the BRDR training setup for solving different problems with PINNs.}	
		\begin{tabular}{cccc}
			\toprule
			Problems           & Allen–Cahn                                                                 & Helmholtz                                                                   & Burgers \\ 
			\midrule
			PDE points         & \begin{tabular}[c]{@{}c@{}}Latin Hypercube\\ 25600\end{tabular}            & \begin{tabular}[c]{@{}c@{}}Uniform\\  101$\times$101\end{tabular}            &  \begin{tabular}[c]{@{}c@{}}Latin Hypercube\\  10000\end{tabular}        \\ \hline
			IC points          & \begin{tabular}[c]{@{}c@{}}Uniform\\  512\end{tabular}                      & --                                                                          &   \begin{tabular}[c]{@{}c@{}}Uniform\\  100\end{tabular}              \\ \hline
			BC points          & --                                                                         & \begin{tabular}[c]{@{}c@{}}Uniform\\  200\end{tabular}                       &    \begin{tabular}[c]{@{}c@{}}Random\\  200\end{tabular}     \\ \hline
			Network            & \begin{tabular}[c]{@{}c@{}}{[}21{]}+{[}128{]}$\times$6+{[}1{]}\\ mFCN \\tanh\end{tabular} & \begin{tabular}[c]{@{}c@{}}{[}2{]}+{[}128{]}$\times$6+{[}1{]}\\ mFCN \\ tanh\end{tabular} & \begin{tabular}[c]{@{}c@{}}{[}2{]}+{[}128{]}$\times$6+{[}1{]}\\ mFCN \\tanh\end{tabular}  \\ \hline
			Adam steps   & 3e5                                                                    & 1e5                  &  4e4   \\ \hline
			Adam Learning rate      & $0.001\times0.99^{n//750}$                                      & $0.005\times0.99^{n//250}$                                       &    $0.001\times0.99^{n//100}$      \\ \hline
			($\beta_c, \beta_w$) in BRDR & (0.999, 0.999)                                                             & (0.999, 0.999)                                                             &  (0.999, 0.999)        \\
			\bottomrule     
		\end{tabular}
		\label{tab_setup_PINN}
	\end{table}

\begin{figure}[htbp]
		\centering	
		\includegraphics{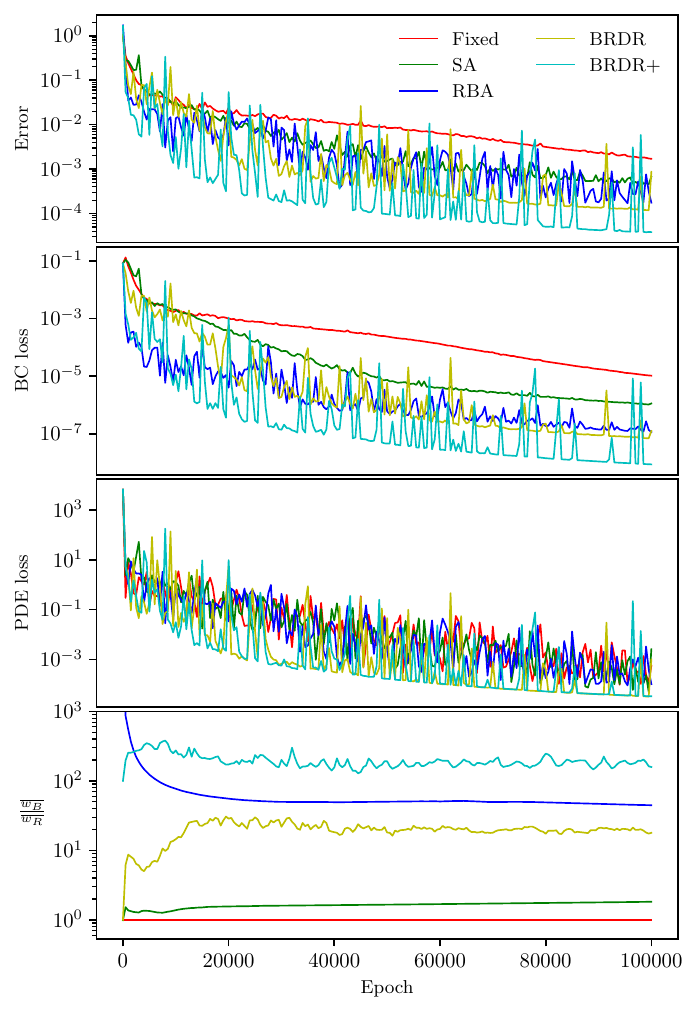}
		\caption{PINN for the 2D Helmholtz equation: The history of $L_2$ relative error, unweighted loss of each component and the average weight ratio of BC to PDE from fixed-weight training, and adaptive-weight training(\enquote{SA}, \enquote{RBA}, \enquote{BRDR}, \enquote{BRDR+}). Note that all the cases share the same network architecture and the same random seed for initialization of network parameters.
		}
		\label{Fig_history_Helmholtz}	
	\end{figure}

	The evolution history of error, loss and the weight ratio of BC to PDE is illustrated in Fig. \ref{Fig_history_Helmholtz}, where the  weight ratio  of BC to PDE is defined as follows:
	\begin{equation}
		\frac{\overline{w_B}}{\overline{w_R}} = \frac{\alpha_B \text{mean}(\mathbf{w}_B)}{\alpha_R\text{mean}(\mathbf{w}_R)}
	\end{equation}
	The error for all the adaptive methods (RBA, SA and BRDR, BRDR+) drops faster than that for fixed weights, highlighting the advantages of adaptive weights. In the first 20,000 epochs, the error for Fixed, SA, and BRDR weights shows a very similar decay rate, all of which are slower than that for RBA weights. This is because RBA manually sets the weights of BC to 100, causing the BC residuals to decay faster initially. This also demonstrates that prediction accuracy is dominated by the BC residual rather than the PDE residuals for this problem. Despite this, the BRDR method gradually catches up with and surpasses the error of RBA as the number of epochs increases, because the average BC weight is rapidly and adaptively increased  at the beginning. Additionally, we can manually set the BC weight constant to $\lambda_B=100$, referred to as BRDR+. With this modification, the error for BRDR+ drops the fastest among all the weighting methods. For example, BRDR+ takes less than half the number of epochs to achieve the final error of RBA. As for the SA weighting method, the weight ratio of BC to PDE converges to about 2 in SA, making the error and BC residuals relatively larger than those of RBA, BRDR, and BRDR+. Since the weights of the SA method are increased proportionally to the square of the residuals and no predefined weight constant is applied to the BC loss, it is difficult for the SA method to achieve a large weight ratio of BC to PDE. This is due to the fact that the BC residuals are much smaller than the PDE residuals. The statistical errors  and computational cost are given in Table \ref{tab_err_PINN}. The computational costs of the adaptive weighting methods (RBA, SA, and BRDR) are very similar, with each being less than 10\% slower than the fixed weighting method. The prediction error of BRDR is almost identical to that of RBA, although no predefined weight constant is used in BRDR. With the predefined weight constant, BRDR+ achieves a much lower prediction error with smaller uncertainty. 
	
	\begin{table}[htbp]
		\centering
		\caption{$L_2$ relative error and relative computational time cost of PINNs  for different weighing methods. The mean and standard deviation are calculated over 5 independent runs. Note that different weighting methods share the same random seed for each run.}
		\begin{tabular}{ccccccc}
			\toprule
			\multirow{2}{*}{\begin{tabular}[c]{@{}c@{}}	Weighting\\methods\end{tabular}}& \multicolumn{2}{c}{2D Helmholtz}                                                  & \multicolumn{2}{c}{1D Allen–Cahn} & \multicolumn{2}{c}{1D Burgers}                        \\  \cmidrule(lr{8pt}){2-3} \cmidrule(lr{8pt}){4-5} \cmidrule(lr{8pt}){6-7}	    & Error & Time & Error & Time & Error & Time               \\			
			\midrule
			Fixed   &   $(2.95\pm0.61)$e-3 &100\% &   $(7.15\pm5.40)$e-4 &100\%  &      $(7.36\pm4.90)$e-4  &100\% \\
			SA \cite{mcclenny2020self} &   $(4.40\pm0.61)$e-4 &102\%     &   $(1.51\pm2.76)$e-4   &101\%      &     $(4.80\pm1.01)$e-4  &103\%  \\ 
			RBA \cite{anagnostopoulos2024residual} &   $(1.95\pm0.20)$e-4   &101\%     &    $(2.92\pm0.78)$e-5 &101\%  &    $(8.22\pm2.33)$e-4 &101\%  \\ 
			BRDR         &    $(1.73\pm0.28)$e-4  &104\%     &    $(2.51\pm0.44)$e-5    &104\%     &  \textbf{$\mathbf{(1.38\pm0.85)}$e-4}   &107\%  \\
			BRDR+         &    \textbf{$\mathbf{(4.86\pm0.18)}$e-5}  &104\%     &    \textbf{$\mathbf{(1.45\pm0.46)}$e-5}   &104\%      &  - & - \\			
			\bottomrule
		\end{tabular}
		\label{tab_err_PINN}
	\end{table}
	
	\FloatBarrier

	\subsection{1D Allen-Cahn equation}\label{sec_AllenCahn}
	The 1D Allen-Cahn equation is defined as follows:
	\begin{equation}\label{eq_AC}
		\begin{aligned}
			\frac{\partial u}{\partial t} -5(u-u^3)-D \frac{\partial^2 u}{\partial x^2} &= 0, \qquad (x,t)\in [-1,1]\times[0,1] \\
			u(x,0)&=x^2 \cos(\pi x), \qquad x\in[-1,1] \\
			u(-1,t)&=u(1,t),        \qquad \qquad t\in[0,1]  \\
		\end{aligned}
	\end{equation}
	and we consider the case of viscosity $D=1E-4.$ 
	
	As adopted in reference \cite{anagnostopoulos2024residual}, we use Fourier feature transformation on $x$ to make the network model automatically satisfy the periodic boundary condition. With 10 Fourier modes, the two-element input $\mathbf{x} = (x, t)$ is lifted to a 21-element input $\widehat{\mathbf{x}}$ before feeding it to the network with the following mapping:
	\begin{equation}
		\label{eq_featureTransform}
		\widehat{\mathbf{x}} = \gamma(\mathbf{x}) =
		\left[
		\sin(\pi \mathbf{B} x),
		\cos(\pi \mathbf{B} x),
		t
		\right]^T, 
	\end{equation}
	where $\mathbf{B} = [1, \ldots, 10]^T$.
	
	The loss function is defined as follows:
	\begin{equation}
		\mathcal{L}(\pmb{\theta}; \mathbf{w}, s, \pmb{\alpha}) = s\left( \frac{\alpha_R}{N_R}\sum_{i=1}^{N_R}w_R^i\mathcal{R}^2(\mathbf{x}_R^i)
		+\frac{\alpha_I}{N_I}\sum_{i=1}^{N_I}w_I^i\mathcal{I}^2(\mathbf{x}_I^i)
		\right)
	\end{equation}
	where $\mathcal{R}$ and $\mathcal{I}$ represent the PDE operator and the initial condition (IC) operator, respectively.
	
	The choice of location of the training points and the BRDR training setup are provided in Table \ref{tab_setup_PINN}. For fixed-weight training, the weights for both the IC and PDE are set to 1.  For the soft-attention (SA) training setup, we use the configuration for Burgers equation from reference \cite{mcclenny2020self}, which lacks tests for the Allen-Cahn equation but is similar.
	The pointwise self-adaptive weights for IC and PDE are all initialized using uniform sampling $\mathcal{U}(0,1)$, and the weights are updated with a fixed learning rate of 0.005. For the residual-based attention (RBA) training setup, we use the configuration provided in reference \cite{anagnostopoulos2024residual}. In this setup, the weights for IC are fixed at 100, and the pointwise self-adaptive weights for PDE are initialized at 0. These weights are then updated with a decay rate of 0.999, an offset of 0  and a learning rate of 0.01. For BRDR+, the weight constant for IC is set as 100.

\begin{figure}[htbp]
		\centering	
		\includegraphics{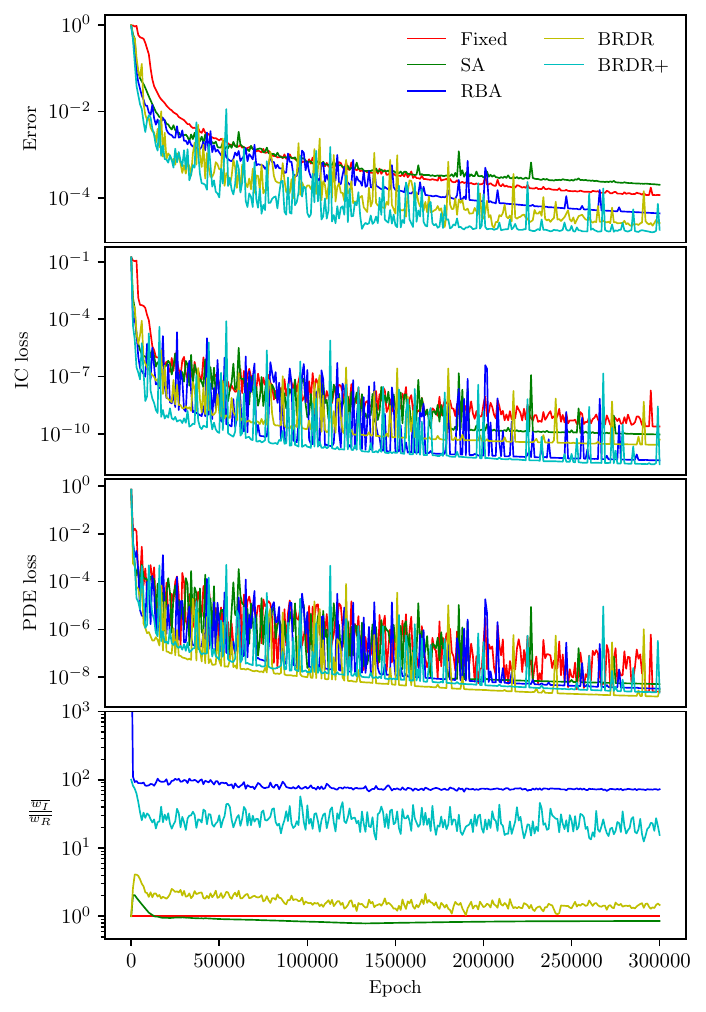}
		\caption{PINN for 1D Allen-Cahn equation: The history of $L_2$ relative error, unweighted loss of each component and the average weight ratio of IC to PDE from fixed-weight training, and adaptive-weight training(\enquote{SA}, \enquote{RBA}, \enquote{BRDR}, \enquote{BRDR+}). Note that all the cases share the same network architecture and the same random seed for initialization of network parameters.
		}
		\label{Fig_history_AllenCahn}	
	\end{figure}

	The evolution history of error, loss, and the weight ratio of initial condition (IC) to partial differential equation (PDE) is illustrated in Fig. \ref{Fig_history_AllenCahn}. The results demonstrate that the error for all adaptive methods decreases more rapidly than for the fixed weight methods, underscoring the advantages of employing adaptive weights. Notably, the error reduction for both BRDR and BRDR+ is significantly faster than that for SA and RBA, particularly in the initial stages of training. Specifically, BRDR achieves the final error level of RBA in less than half the number of epochs, while BRDR+ achieves the same error level in less than one third of the epochs.
	In terms of weight allocation, BRDR+ assigns more weight to the IC, resulting in the smallest IC loss among all the weighting methods at the end of training. In contrast to BRDR+, BRDR assigns more weight to the PDE, leading to the smallest PDE loss at the end of training. The fact that the prediction error of BRDR+ is smaller than that of BRDR suggests that, for this particular problem, prediction accuracy is more heavily influenced by the residuals of the IC rather than those of the PDE. Without a predefined weight constant, the weight ratio of IC to PDE in SA falls below 1, thereby failing to adequately recognize the importance of the IC.
	Furthermore, the statistical errors associated with each method are provided in Table \ref{tab_err_PINN}. Both BRDR and BRDR+ exhibit significant improvements over SA and RBA in terms of both the magnitude and uncertainty of the prediction error. This highlights the efficacy of the BRDR and BRDR+ methods in enhancing the accuracy and reliability of the predictions in the context of adaptive weighting schemes.

	Given the numerous components employed in the test cases, we isolated each component to evaluate its individual impact. Accordingly, an ablation study was conducted on the Allen-Cahn equation to analyze the contributions of the modified fully-connected network, Fourier feature embedding, the scaling factor in the BRDR method, and the pointwise weights in the BRDR method. The results, presented in Section \ref{sec_ablation}, demonstrate that each component individually contributes to error reduction, with the extent of that reduction varying by component.

	\FloatBarrier
	
	\subsection{1D Burgers equation}
	\label{sec_Burgers}
	The 1D Burgers equation is defined as follows:
	\begin{equation}
		\begin{aligned}
			\frac{\partial u}{\partial t} + u\frac{\partial u}{\partial x} -\nu \frac{\partial^2 u}{\partial x^2} &= 0, \qquad (x,t)\in [-1,1]\times[0,1] \\
			u(x,0)&=-\sin(\pi x), \qquad x\in[-1,1] \\
			u(\pm 1,t)&=0,        \qquad \qquad t\in[0,1]  \\
		\end{aligned}
	\end{equation}
	where $u$ is the flow velocity, and we consider the case with viscosity $\nu=0.01/\pi.$

	The loss function is defined as follows:
	\begin{equation}
		\mathcal{L}(\pmb{\theta}; \mathbf{w},s,\pmb{\alpha}) = s \left( \frac{\alpha_R}{N_R}\sum_{i=1}^{N_R}w_R^i\mathcal{R}^2(\mathbf{x}_R^i)
		+\frac{\alpha_B}{N_B}\sum_{i=1}^{N_B}w_B^i\mathcal{B}^2(\mathbf{x}_R^i)
		+\frac{\alpha_I}{N_I}\sum_{i=1}^{N_I}w_I^i\mathcal{I}^2(\mathbf{x}_R^i)
		\right)
	\end{equation}
	where $\mathcal{R}$, $\mathcal{B}$ and $\mathcal{I}$ represent the PDE operator, the BC operator and the IC operator, respectively.
	
	The choice of location of the training points and the BRDR training setup are provided in Table \ref{tab_setup_PINN}. For fixed-weight training, the weights for the IC, BC and PDE points are set to 1. For the soft-attention (SA) training setup, we follow the configuration given in reference \cite{mcclenny2020self} for Adam training of Burgers equation. The pointwise self-adaptive weights for the IC, BC and PDE points are all initialized using uniform sampling $\mathcal{U}(0,1)$, and the weights are updated with a fixed learning rate of 0.005.
	Since Burgers equation is not tested with RBA weights in reference \cite{anagnostopoulos2024residual}, we set the weights for BC and IC to 1, and the pointwise self-adaptive weights for PDE are initialized at 0. These weights are then updated with a decay rate of 0.999, an offset of 0, and a learning rate of 0.01. As we have not found specific weight constants for BRDR+ to surpass BRDR, we only provide a comparison only for the fixed weight, SA, RBA, and BRDR setups.

\begin{figure}[htbp]
		\centering	
		\includegraphics{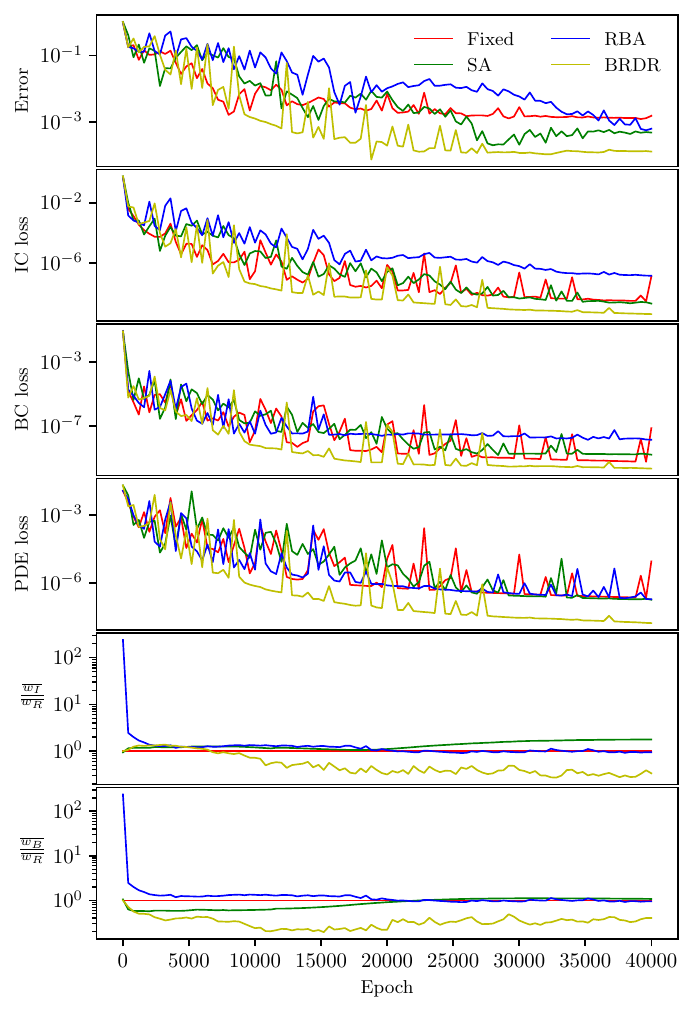}
		\caption{PINN for 1D Burgers equation: The history of $L_2$ relative error, unweighted loss of each component, the average weight ratio of IC to PDE, the average weight ratio of BC to PDE from fixed-weight training, and adaptive-weight training(\enquote{SA}, \enquote{RBA}, \enquote{BRDR}). Note that all the cases share the same network architecture and the same random seed for initialization of network parameters.
		}
		\label{Fig_history_Burgers}	
	\end{figure}

	The evolution history of error, loss, and weight ratios is illustrated in Fig. \ref{Fig_history_Burgers}. The results demonstrate that the error for all adaptive methods (RBA, SA, and BRDR) decreases more rapidly compared to the fixed weight methods, highlighting the advantages of adaptive weighting methods. Notably, the error reduction for BRDR is significantly faster than that for SA and RBA, and this trend is similarly observed in the IC loss, BC loss, and PDE loss. Specifically, BRDR achieves the final error level of RBA and SA in less than half the number of epochs.
	The plots of weight ratios reveal that both SA and RBA assign more weight to the IC and BC, whereas BRDR allocates more weight to the PDE. Despite this, the IC, BC, and PDE losses for BRDR are smaller than those for SA and RBA, underscoring the high convergence rate of the BRDR method. The statistical errors associated with each method are provided in Table \ref{tab_err_PINN}. The average error of RBA is slightly larger than that of fixed weights, as its adaptive weights focus on the large gradient part of the domain, which has not been resolved. This observation is also reported in \cite{anagnostopoulos2024learning}. BRDR significantly outperforms SA and RBA in terms of both the magnitude and uncertainty of the prediction error.
	
	\FloatBarrier

	\subsection{Summary}\label{sec_PINN_summary}
	As demonstrated in the three benchmarks, BRDR exhibits higher accuracy, convergence rate and lower uncertainty. Additionally, compared to the RBA method, BRDR weights can be applied to all training points within a unified framework (similar to the SA method), eliminating the need to manually set weights for IC or BC components.
	Manually choosing weights often requires extensive hyperparameter tuning, which is labor-intensive. However, if a suitable set of weight constants is available, the performance of BRDR can be further improved. Compared to the SA method, BRDR weights are bounded (similar to the RBA method), which prevents issues with weight explosion during updates.
	In the three benchmarks, we consistently used the same BRDR hyperparameters, $(\beta_c, \beta_w) = (0.999, 0.999)$. Based on our testing experience, setting $\beta_c$ or $\beta_w$ to 0.999 or 0.9999 is sufficient for fast training. Consequently, BRDR could significantly reduce the labor involved in hyperparameter tuning.

\begin{figure}[htbp]
		\centering	
		\includegraphics[ width=14cm, trim=0cm 11.0cm 0cm 0cm, clip=true]{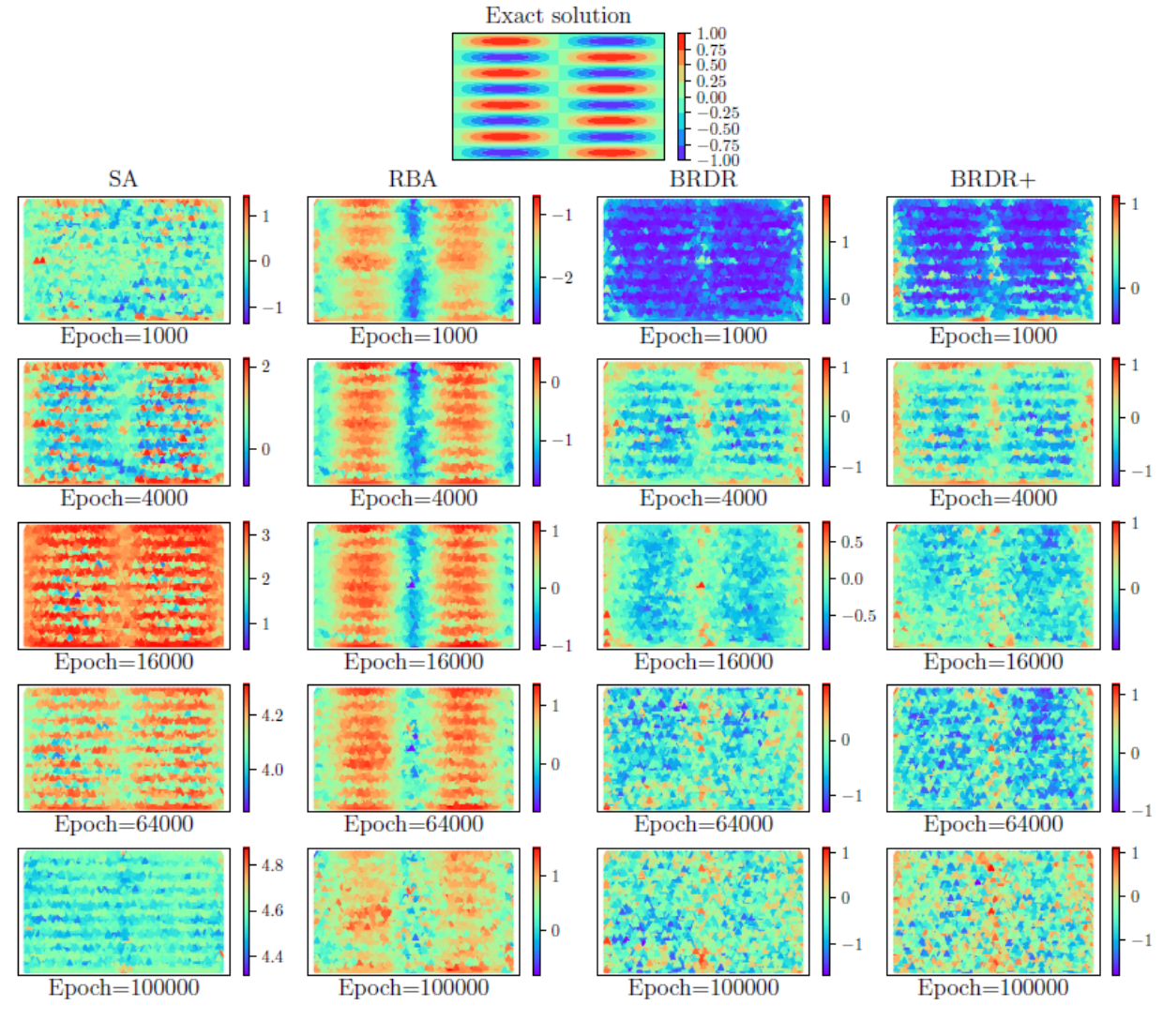}
		\caption{PINN for 2D Helmholtz equation: the exact solution (top middle) and the evolution history of the distribution  of adaptive weights ($\log_{10}{w}$) for  adaptive-weight training(\enquote{SA}, \enquote{RBA}, \enquote{BRDR}, \enquote{BRDR+}). Note that all the cases share the same network architecture and the same random seed for initialization of network parameters. }
		\label{Fig_SAweightsHelmholtz}	
	\end{figure}

\begin{figure}[htbp]
		\centering	
		\includegraphics[ width=14cm, trim=0cm 11.5cm 0cm 0cm, clip=true]{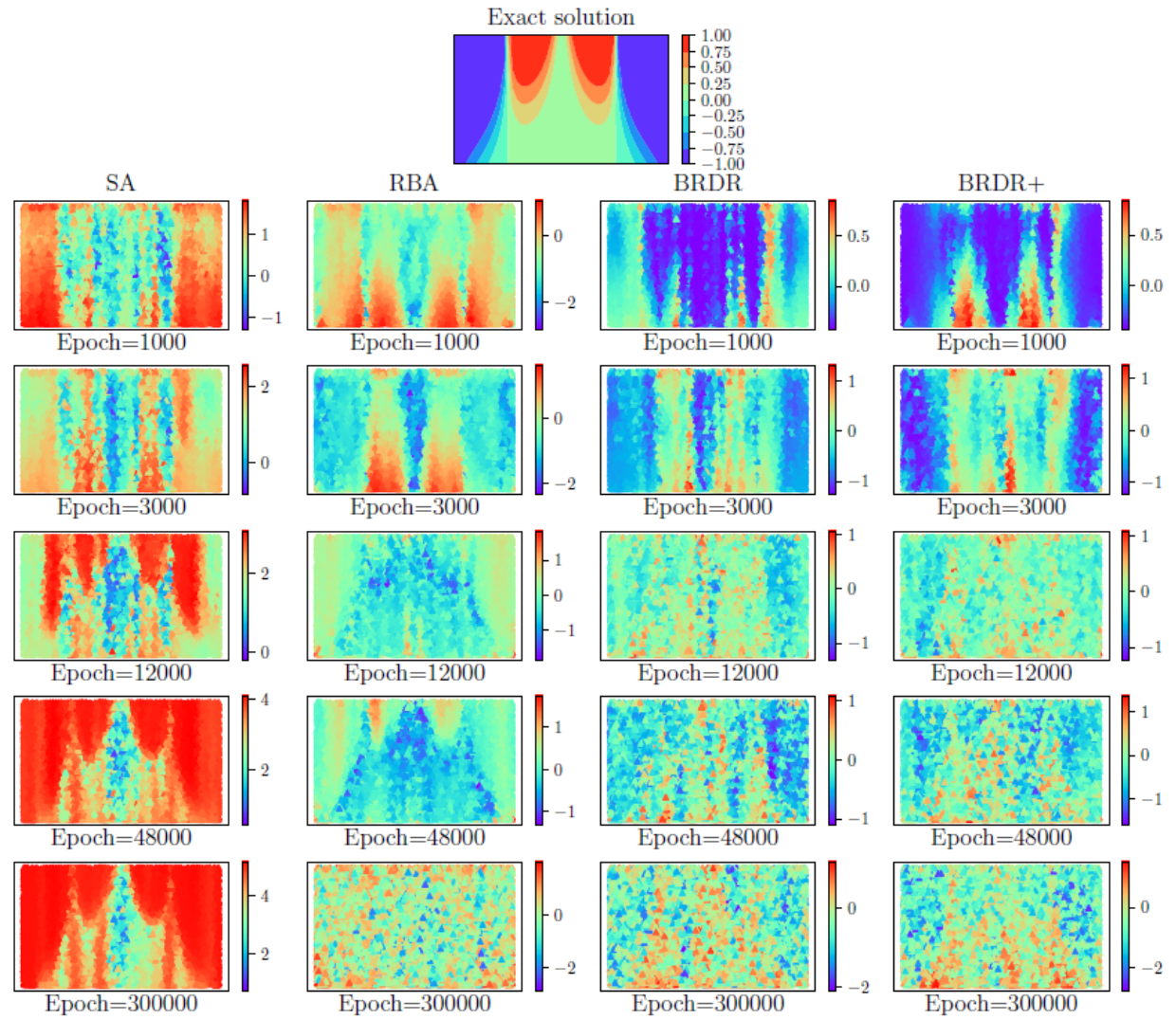}
		\caption{PINN for 1D Allen-Cahn equation: the evolution history of the distribution  of adaptive weights ($\log_{10}{w}$) at PDE training points for  adaptive-weight training(\enquote{SA}, \enquote{RBA}, \enquote{BRDR}, \enquote{BRDR+}). Note that all the cases share the same network architecture and the same random seed for initialization of network parameters. }
		\label{Fig_SAweightsAllenCahn}	
	\end{figure}

\begin{figure}[htbp]
		\centering	
		\includegraphics[ width=12cm, trim=1cm 5.5cm 1cm 0cm]{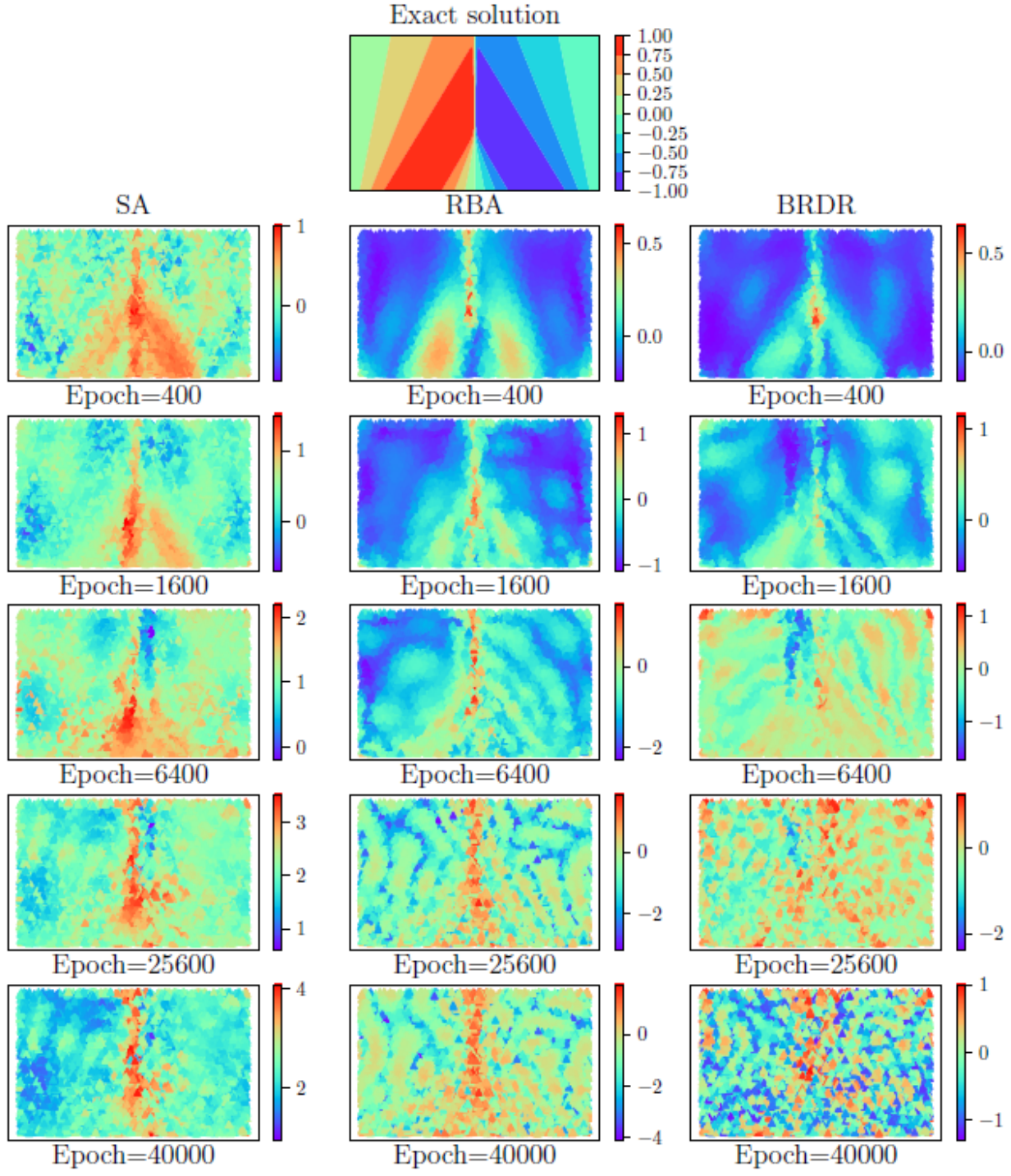}
		\caption{PINN for 1D Burgers equation: the exact solution (top middle) and the exact solution (top middle) and the evolution history of the distribution  of adaptive weights ($\log_{10}{w}$) at PDE training points for  adaptive-weight training(\enquote{SA}, \enquote{RBA}, \enquote{BRDR}). Note that all the cases share the same network architecture and the same random seed for initialization of network parameters. }
		\label{Fig_SAweightsBurgers}	
	\end{figure}

	Additionally, the evolution history of adaptive weights at PDE training points is illustrated in Figs. \ref{Fig_SAweightsHelmholtz}, \ref{Fig_SAweightsAllenCahn}, and \ref{Fig_SAweightsBurgers}. For most test cases, the adaptive weights initially exhibit low-frequency, large-scaling features that correspond to the overall structure of the solutions. As the number of epochs increases, the weight distribution transitions to higher frequencies and becomes more homogeneous. This phenomenon is also reported in \cite{anagnostopoulos2024residual}. This evolution aligns with the dynamics of the training process, wherein the training initially resolves low-frequency, large-scale modes, and subsequently addresses high-frequency, smaller-scale structures.
	We suspect that if adaptive weights exhibit a distinctive structure, it indicates that the training with adaptive weights is focusing on resolving the corresponding scale structure in the solution. As the corresponding scale structure is resolved, the weight distribution will transition to smaller-scale structures to address finer details. Therefore, a more homogeneous distribution of weights suggests that the solution has been better resolved.
	However, obvious non-homogeneity is observed in some cases, such as the SA weight distribution for the Allen-Cahn equation in Fig. \ref{Fig_SAweightsAllenCahn}, the SA weight distribution for the Burgers equation in Fig. \ref{Fig_SAweightsBurgers}, and the RBA weight distribution for the Burgers equation in Fig. \ref{Fig_SAweightsBurgers}. As a result, the corresponding error is relatively larger.
	
	By conducting a more detailed analysis of the Burgers equation, we observe that both the SA and RBA methods assign larger weights near the viscous shock region. This may explain their inferior performance compared to the BRDR method. The key challenge in solving the Burgers equation with small viscosity is that we are essentially attempting to handle a discontinuous solution using a differential formulation. A recent study \cite{liu2024discontinuity} has shown that assigning lower weights to regions characterized by steep gradients or discontinuities can yield impressive results for shock problems.
		This insight runs counter to the RBA and SA weighting strategies, which naturally allocate more weight to points with larger residuals. As a result, these methods may overemphasize the challenging, discontinuous regions, thereby hindering their overall convergence efficiency. In contrast, our approach adapts the weights based on the residual decay rate rather than the residual magnitude. Consequently, even if the residual is large near the shock, its decay rate may not be significantly lower than in other regions. This prevents our method from disproportionately focusing on the discontinuity.
		As illustrated in Fig. \ref{Fig_SAweightsBurgers}, our weight distribution remains more uniform, suggesting a more balanced training process. This provides a possible explanation for the improved performance of our method, as it avoids the pitfall of over-allocating computational effort to the most challenging parts of the domain.

	\section{Numerical results for physics-informed operator learning}
	\label{sec_test_DeepOnet}
	To further validate the performance of the BRDR weighting method, we applied it to training physics-informed deep operator networks (PIDeepONets) \cite{wang2021learning,goswami2022physics,goswami2023physics}. We have studied its performance for for two operator learning problems, for the 1D wave equation and the 1D Burgers equation. In this setup, PIDeepONets are employed to learn the solution $G_{\pmb{\theta}}(\mathbf{u}_0)(\mathbf{x})$ with respect to coordinates $\mathbf{x} = (x,t)$ corresponding to the initial condition  $\mathbf{u}_0 = u_0(\mathbf{x})$.  In the previous section on PINN training, we compared our method with soft-attention (SA) \cite{mcclenny2020self}, residual-based attention (RBA) \cite{anagnostopoulos2024residual}, and fixed weights methods. However, since SA and RBA are not specifically designed for PIDeepONets, in this section, we compare our method with two specific weighting methods tailored for PIDeepONet: the Neural Tangent Kernel (NTK) weighting method \cite{wang2022improved}, the Conjugate Kernel (CK) weighting method \cite{howard2024conjugate}, as well as the fixed weights method. The reported error is defined as the average $L_2$ relative error:
	\begin{equation}
		\epsilon_{L_2} = \frac{1}{N} \sum_{i=1}^{N} \frac{\|u(\mathbf{x}; \mathbf{u}_0^i) - u_E(\mathbf{x}; \mathbf{u}_0^i)\|_2}{\|u_E(\mathbf{x}; \mathbf{u}_0^i)\|_2}
	\end{equation}
	where $N$ is the number of test instances, and $u(\mathbf{x}; \mathbf{u}_0^i)$ and $u_E(\mathbf{x}; \mathbf{u}_0^i)$ are vectors of the predicted solutions and the exact solutions given the initial condition $\mathbf{u}_0^i$, respectively.

	In this section, we use the mDeepONet network architecture (see \ref{sec_net}), where both the trunk and branch networks are built with 7 hidden layers, each containing 100 neurons. The hyperbolic tangent function is employed as the activation function. The network parameters are initialized using the Kaiming Uniform initialization \cite{he2015delving}. Specifically, for a module of shape (out\_features, in\_features), the learnable weights and biases are initialized from $\mathcal{U}(-\sqrt{k}, \sqrt{k})$, where $k = 1/\text{in\_features}$.
	We use only the Adam optimizer \cite{kingma2014adam} for updating the training parameters with a mini-batch training strategy. The batch size is set to 10,000, and the learning rate is initialized at 0.001, decaying by 0.99 every 500 steps. The hyperparameters in the BRDR weighting method are set to $(\beta_c, \beta_w) = (0.9999, 0.999)$.
	All training procedures described in this section are implemented using PyTorch \cite{paszke2019pytorch}. Training computations were performed on a GPU cluster, with each individual training run utilizing a single NVIDIA\textsuperscript{\textregistered} Tesla P100 GPU. All computations were conducted using 32-bit single-precision floating-point format.
	
	\subsection{1D Wave equation}\label{sec_DeepONet_Wave}
	The 1D wave equation is defined as:
	\begin{align}
		\frac{\partial^2 u}{\partial t^2}-C^2\frac{\partial^2 u}{\partial x^2}&= 0, \qquad &(x,t)\in [0,1]^2 \\
		u(x,0)&=u_0(x), \qquad &x\in[0,1] \\
		\frac{\partial{u}}{\partial{t}}(x,0)&=0, \qquad &x\in[0,1] \\
		u(0,t)=u(1,t)&=0, \qquad &t\in[0,1]
	\end{align}
	and we consider the case where the wave velocity is $C=\sqrt{2}.$ The initial condition is set to $u_0(x) = \sum_{n=1}^{5} b_n \sin(n\pi x)$. The exact solution is given by $u(x,t; u_0) = \sum_{n=1}^{5} b_n \sin(n\pi x) \cos(n\pi C t)$. For training the PIDeepONet, 1000 random initial conditions, each represented by 101 uniform \(x\) points, are generated by randomly sampling $\{b_n\}_{n=1}^{5}$ from the normalized Gaussian distribution.
	For each initial condition, 100 boundary points are randomly sampled on the boundaries $x = 0$ and $x = 1$, and 2500 residual points are randomly sampled in the domain $(x,t) \in [0,1]^2$. For testing, 500 random initial conditions are sampled, each represented by 101 uniform $x$ points. For these initial conditions, the solution values at $101 \times 101$ uniformly sampled spatiotemporal points  are computed using the exact solution.

	The loss function is defined as follows:
	\begin{equation}
		\begin{aligned}
			\mathcal{L}(\pmb{\theta}; \mathbf{w},s) = s \bigg( &\frac{1}{N_R}\sum_{k=1}^{N_{Rb}}w_R^{i_k}\mathcal{R}^2(\mathbf{x}_R^{i_k})
			+\frac{1}{N_B}\sum_{k=1}^{N_{Bb}}w_I^{i_k}\mathcal{B}^2(\mathbf{x}_B^{i_k}) \\
			+&\frac{1}{N_I}\sum_{k=1}^{N_{Ib}}w_I^{i_k}\mathcal{I}^2(\mathbf{x}_I^{i_k})
			+\frac{1}{N_I}\sum_{k=1}^{N_{Ib}}w_{I_t}^{i_k}\mathcal{I}_t^2(\mathbf{x}_I^{i_k})
			\bigg)
		\end{aligned}
	\end{equation}
	where $\mathcal{R}$,  $\mathcal{B}$, $\mathcal{I}$ and $\mathcal{I}_t$  represent the PDE operator, the boundary condition, the zero-order initial condition (IC) and the first-order initial condition (IC\_t) operator, respectively. $\{i_k\}_{k=1}^{N_{Rb}}$, $\{i_k\}_{k=1}^{N_{Ib}}$ and $\{i_k\}_{k=1}^{N_{Bb}}$ are batch indexes of training points, where
	$N_{Rb}=N_{Bb}=N_{Ib}=10000$.
	
\begin{figure}[htbp]
		\centering	
		\includegraphics{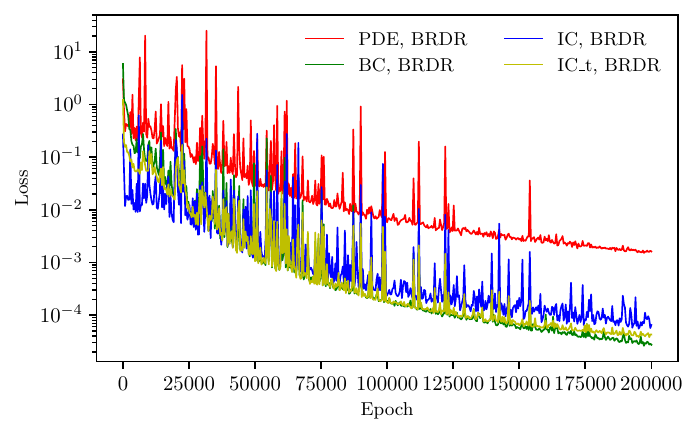}
		\caption{PIDeepONet for the 1D wave equation: the history of unweighted loss of each component from BRDR training. 
		}
		\label{Fig_history_WaveOperator}	
	\end{figure}

\begin{figure}[htbp]
		\centering	
		\includegraphics[ width=12cm, trim=1cm 0.5cm 1cm 0.5cm, clip=true]{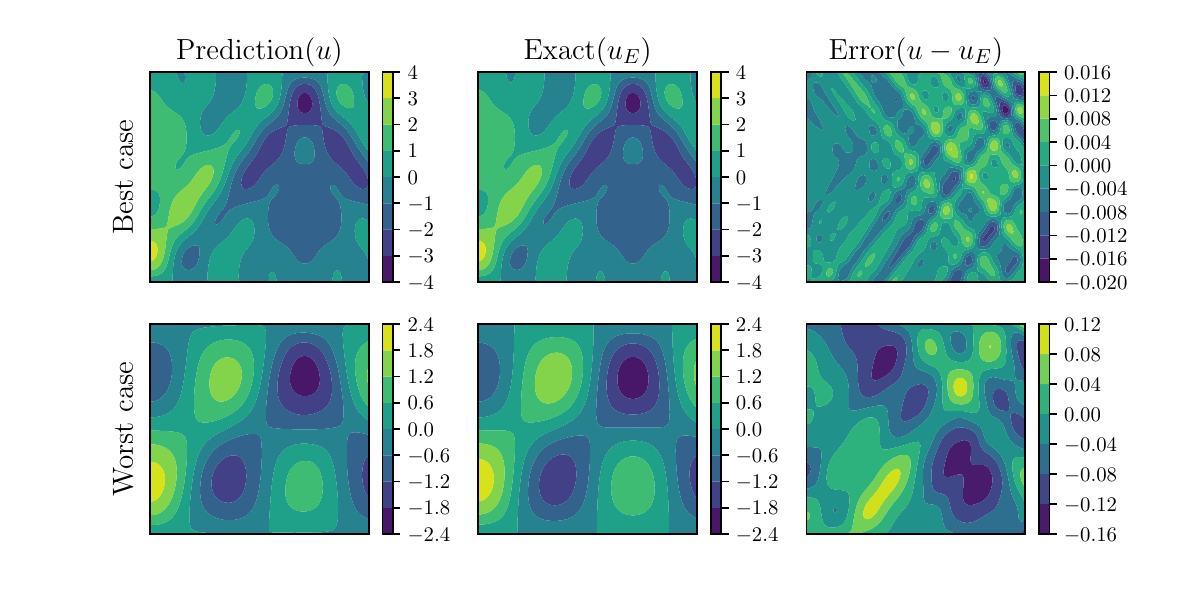}
		\caption{PIDeepONet for the 1D wave equation: the worst and best predicting cases in the test set from BRDR training method. }
		\label{Fig_solutionWaveOperator}	
	\end{figure}

	The loss history of each component is illustrated in Fig. \ref{Fig_history_WaveOperator}. Among all the loss components, the PDE loss exhibits the slowest decay, creating a bottleneck in the training process. Fig. \ref{Fig_solutionWaveOperator} presents the best and worst predictions in the test set. The error distribution clearly follows the two characteristic directions $ x \pm Ct = 0 $, indicating that the BRDR training method is effectively capturing the characteristic structure. However, the non-homogeneous error distribution suggests that BRDR training has not yet fully resolved the PDE, and more epochs are required for further training.
	The prediction errors are detailed in Table \ref{tab_err_PIDeepONet}. The errors associated with all adaptive methods (NTK, CK, and BRDR) are smaller than those with fixed weights, underscoring the benefits of adaptive weighting. The prediction error of BRDR method is comparable with those of NTK and CK  methods. To further examine the distribution of prediction errors across different initial conditions, we also present box plots of the prediction errors from various runs in Fig. \ref{Fig_ErrorBox_BurgersOperator}(a). These plots demonstrate that the BRDR method consistently achieves superior uniformity in prediction errors compared to both NTK and CK methods. Although the mean prediction error of CK is lower than that of BRDR, the extent of outliers is less pronounced with the BRDR method. We believe this increased uniformity is attributed to the inherent uniformity in the convergence rate of the BRDR method.
	Furthermore, BRDR training demonstrates considerable advantages in terms of computational time cost, as shown in Table \ref{tab_err_PIDeepONet}. The NTK weighting method, for instance, involves the evaluation of the NTK matrix, which is highly computationally expensive, requiring approximately 3-4 times more computational time than fixed weight training. The CK method uses an inexpensive approximation of the NTK matrix with little sacrifice in accuracy, though it still incurs a higher extra cost compared to BRDR. In contrast, BRDR training incurs less than a 10\% additional cost, making it a more efficient method.

	\begin{table}[tbp]
		\centering
		\caption{Relative error and relative consumed time for operator learning of the 1D wave equation and 1D Burgers equation. The mean and standard deviation are calculated over 5 independent runs.}
		\begin{tabular}{ccccccc}
			\hline
			& \multicolumn{4}{c}{Burgers}                                                  & \multicolumn{2}{c}{Wave}                       \\  \cmidrule(lr{8pt}){2-5} \cmidrule(lr{8pt}){6-7}
			& \multicolumn{3}{c}{Error}                            & \multirow{2}{*}{Time} & \multirow{2}{*}{Error} & \multirow{2}{*}{Time} \\ \cmidrule(lr{8pt}){2-4}
			& $\nu$=1e-2          & $\nu$=1e-3          & $\nu$=1e-4           &                       &                        &                       \\
			\midrule
			Fixed & 3.18\%$\pm$0.49\% & 8.43\%$\pm$0.87\% & 23.39\%$\pm$1.14\% & 100\%                 & 2.84\%$\pm$0.63\%        & 100\%                 \\
			NTK \cite{wang2022improved}   & 1.04\%$\pm$0.25\% & \textbf{3.15\%$\pm$0.39\%} & \textbf{11.18\%$\pm$1.08\%} & 385\%                 & 1.43\%$\pm$0.42\%        & 456\%                 \\
			CK \cite{howard2024conjugate}     & 0.74\%$\pm$0.10\% & 3.42\%$\pm$0.47\% & 16.85\%$\pm$2.86\%  & 142\%                 & \textbf{0.72\%$\pm$0.04}\%        & 144\%                 \\
			BRDR   & \textbf{0.26\%$\pm$0.01\%} & 3.40\%$\pm$ 0.07\% & 17.61\%$\pm$0.67\% & 106\%                 & 0.92\%$\pm$0.21\%        & 106\%   \\
			\bottomrule             
		\end{tabular}
		\label{tab_err_PIDeepONet}
	\end{table}
	
	\subsection{1D Burgers equation}
	The Burgers equation is defined as:
	\begin{align}
		\frac{\partial u}{\partial t} + u\frac{\partial u}{\partial x} -\nu \frac{\partial^2 u}{\partial x^2} &= 0, \qquad &(x,t)\in [0,1]^2, \\
		u(x,0)&=u_0(x), \qquad &x\in[0,1], \\
		u(0,t)&=u(1,t),        \qquad \qquad &t\in[0,1],  \\
		\frac{\partial{u}}{\partial{x}}(0,t) &= \frac{\partial{u}}{\partial{x}}(1,t), \qquad \qquad &t\in[0,1], 
	\end{align}
	where $\nu$ is the viscosity. $u(x,t; u_0)$ is the solution at the point $\mathbf{x}=(x,t)$  given the initial condition $u_0(x)$. 
	According to reference \cite{wang2022improved}, the initial condition, $u_0(x)$, is sampled from the Gaussian random field $\mathcal{N}(0, 25^2(-\Delta + 5^2I)^{-4})$. For training, 1000 random initial conditions are sampled, each represented by 101 random $x$ points. For each initial condition, 100 boundary points are randomly sampled on the boundaries $x = 0$ and $x = 1$, and 2500 residual points are randomly sampled in the domain $(x, t) \in [0, 1]^2$. For testing, 500 random initial conditions are sampled, each represented by 101 uniform $x$ points. For these initial conditions, the solutions at $101 \times 101$ uniformly sampled spatiotemporal points  are computed using the Chebfun package \cite{driscoll2014chebfun}, with the Fourier method for spatial discretization and a fourth-order stiff time-stepping scheme for marching in time.

	The loss function is defined as follows:
	\begin{equation}
		\begin{aligned}
			\mathcal{L}(\pmb{\theta}; \mathbf{w},s) = s \bigg( &\frac{1}{N_R}\sum_{k=1}^{N_{Rb}}w_R^{i_k}\mathcal{R}^2(\mathbf{x}_R^{i_k})
			+\frac{1}{N_I}\sum_{k=1}^{N_{Ib}}w_I^{i_k}\mathcal{I}^2(\mathbf{x}_I^{i_k}) \\
			+&\frac{1}{N_B}\sum_{k=1}^{N_{Bb}}w_B^{i_k}\mathcal{B}^2(\mathbf{x}_B^{i_k})
			+\frac{1}{N_B}\sum_{k=1}^{N_{Bb}}w_{B_x}^{i_k}\mathcal{B}_x^2(\mathbf{x}_B^{i_k})
			\bigg)
		\end{aligned}
	\end{equation}
	where $\mathcal{R}$, $\mathcal{I}$, $\mathcal{B}$ and $\mathcal{B}_x$  represent the PDE operator, the initial condition (IC), zero-order boundary condition (BC) and first-order boundary condition (BC\_x) operator, respectively. $\{i_k\}_{k=1}^{N_{Rb}}$, $\{i_k\}_{k=1}^{N_{Ib}}$ and $\{i_k\}_{k=1}^{N_{Bb}}$ are batch indexes of training points, where
	$N_{Rb}=N_{Bb}=N_{Ib}=10000$.

\begin{figure}[!ht]
		\centering	
		\includegraphics{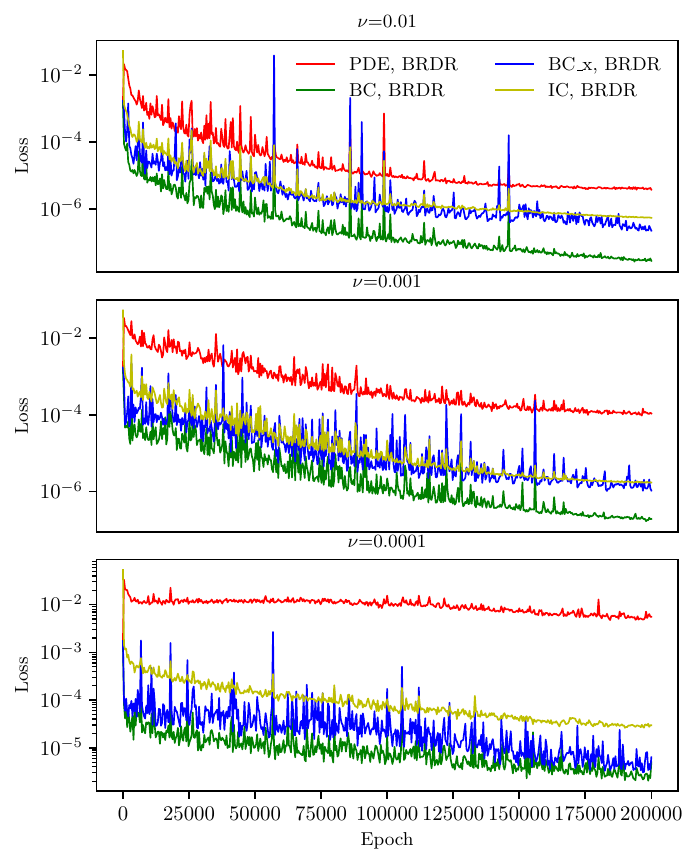}
		\caption{PIDeepONets for 1D Burgers equation: the history of unweighted loss of each component from BRDR training. 
		}
		\label{Fig_history_BurgersOperator}	
	\end{figure}

\begin{figure}[htbp]
		\centering	
		\subfigure{
			\includegraphics[ width=12cm, trim=1cm 0.5cm 1cm 0.5cm, clip=true]{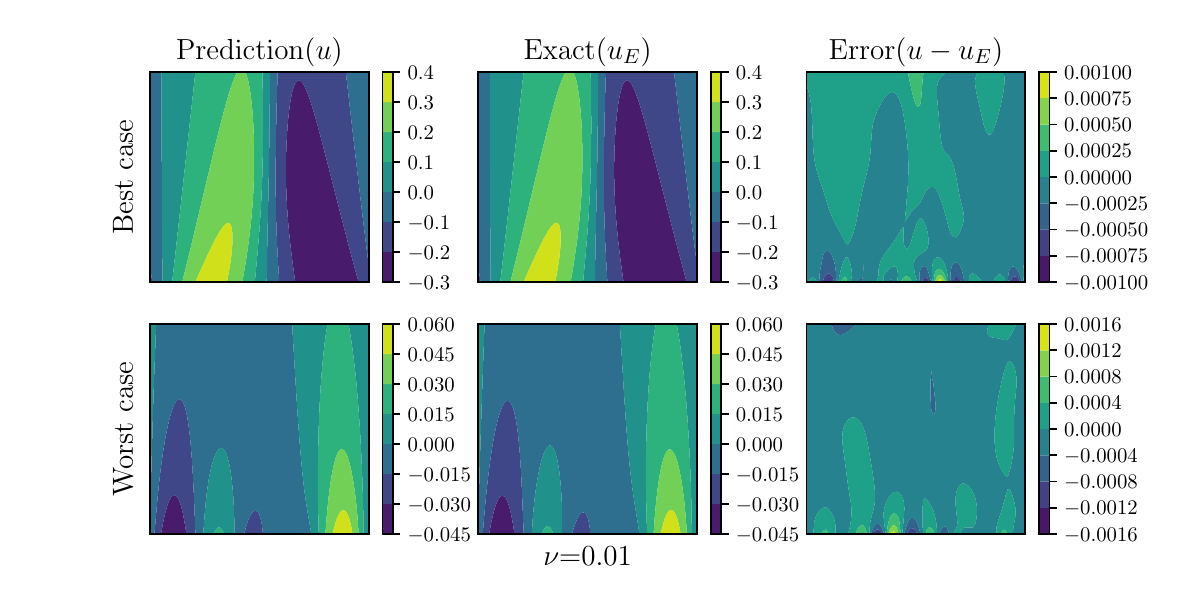}
		}
		\subfigure{
			\includegraphics[ width=12cm, trim=1cm 0.5cm 1cm 0.5cm, clip=true]{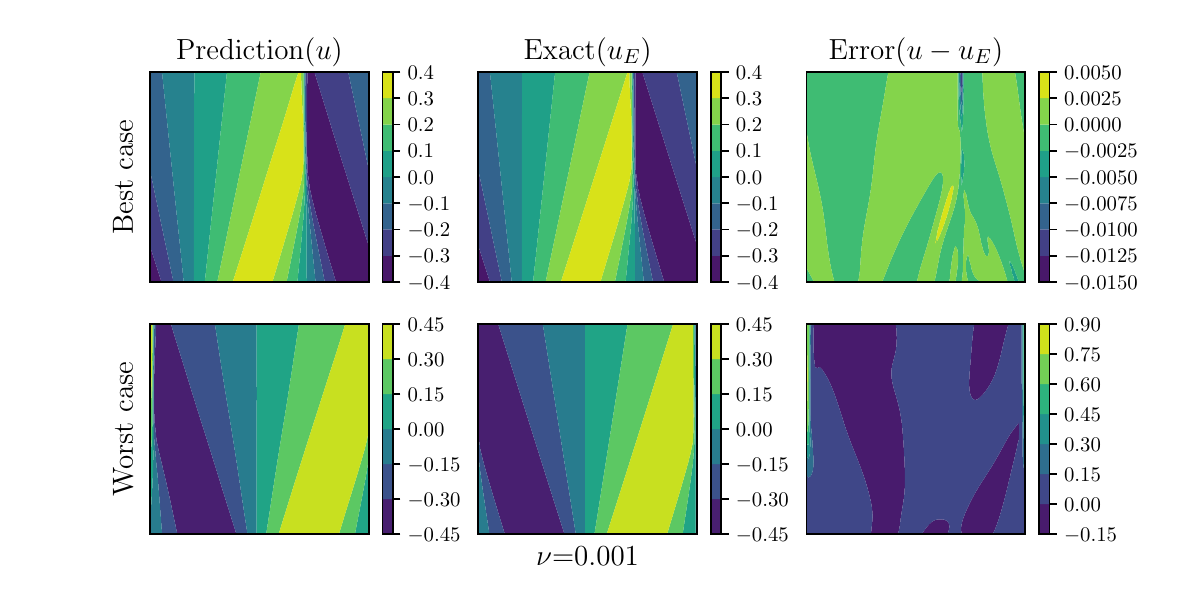}
		}
		\subfigure{
			\includegraphics[ width=12cm, trim=1cm 0.5cm 1cm 0.5cm, clip=true]{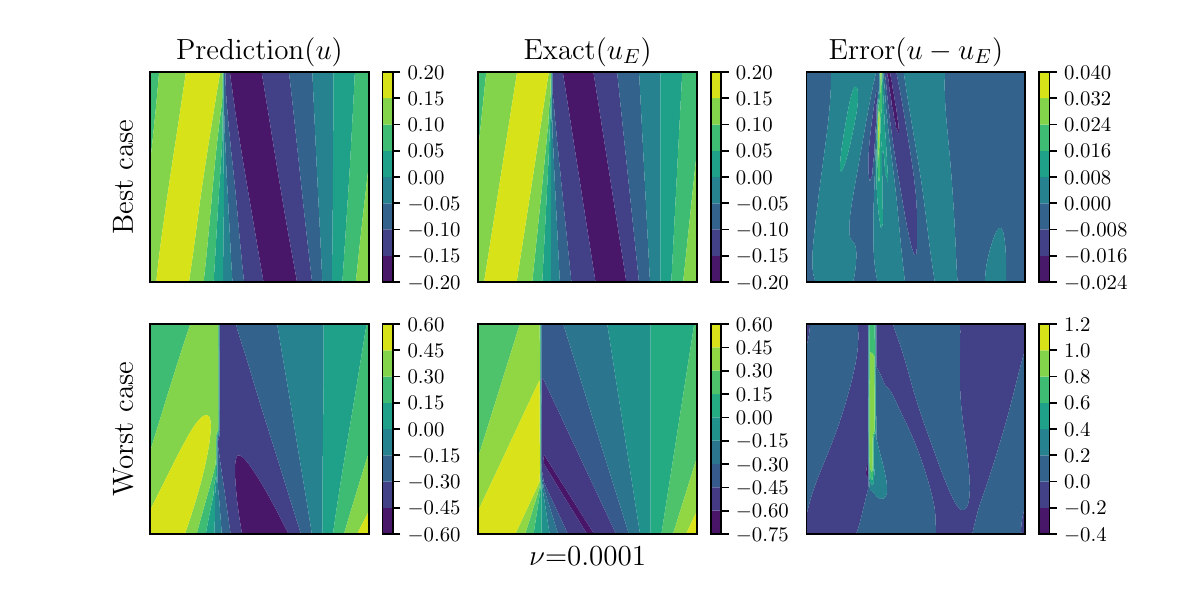}
		}
		\caption{PIDeepONets for 1D Burgers equation: the worst and best predicting cases from BRDR weighting method for different viscosity. }
		\label{Fig_solutionBurgersOperator}	
	\end{figure}

	The loss history of each component is illustrated in Fig. \ref{Fig_history_BurgersOperator}. As the viscosity $\nu$ decreases, the training becomes increasingly difficult, as evidenced by the progressively slower convergence rate of all loss components. Among these, the PDE loss exhibits the slowest decay, creating a bottleneck in the training process. Fig. \ref{Fig_solutionBurgersOperator} displays the best and worst predictions in the test set. For $\nu = 0.01$, the largest error is primarily located at the initial boundary, and the steep gradient is not particularly pronounced, making the training relatively straightforward. For $\nu = 0.001$ and $\nu = 0.0001$, the steep gradient becomes more pronounced, posing a more significant challenge. In the worst cases for $\nu = 0.001$ and $\nu = 0.0001$, the predictions closely align with the ground truth, but there is a slight shift in the steep gradient location prediction, resulting in large errors around it and relatively smaller errors in the smoother areas of the solution. This suggests that adaptive sampling is necessary to further refine the steep gradient area.
	
\begin{figure}[!ht]
		\centering	
		\subfigure{
			\includegraphics[ width=14cm, trim=1cm 0.3cm 1.0cm 12cm, clip=true]{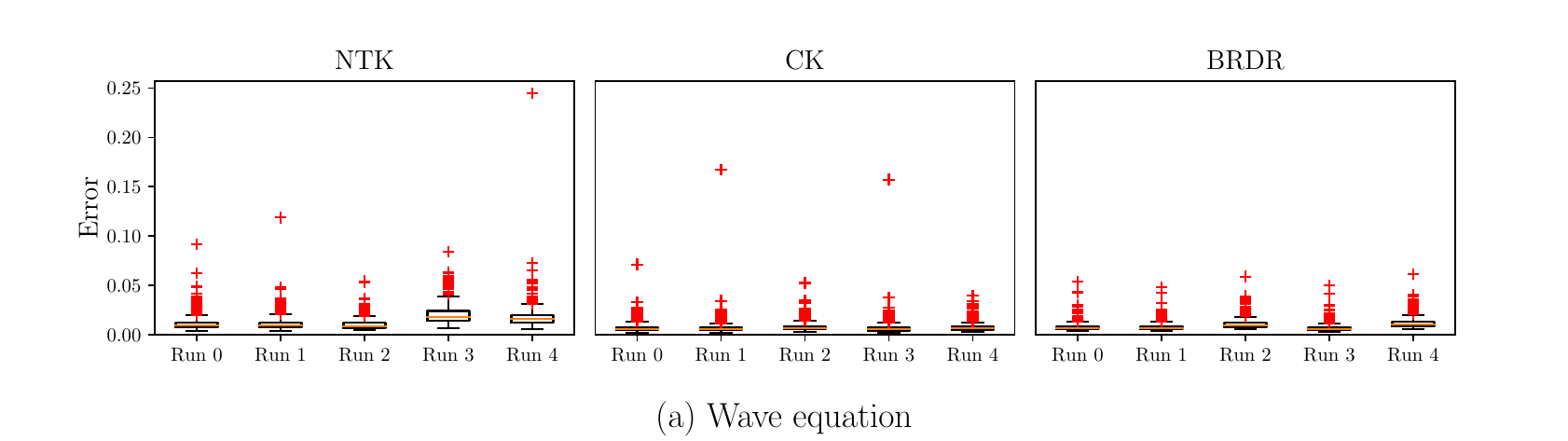}
		} 
		\subfigure{
			\includegraphics[ width=14cm, trim=1cm 0.3cm 1.0cm 1.8cm, clip=true]{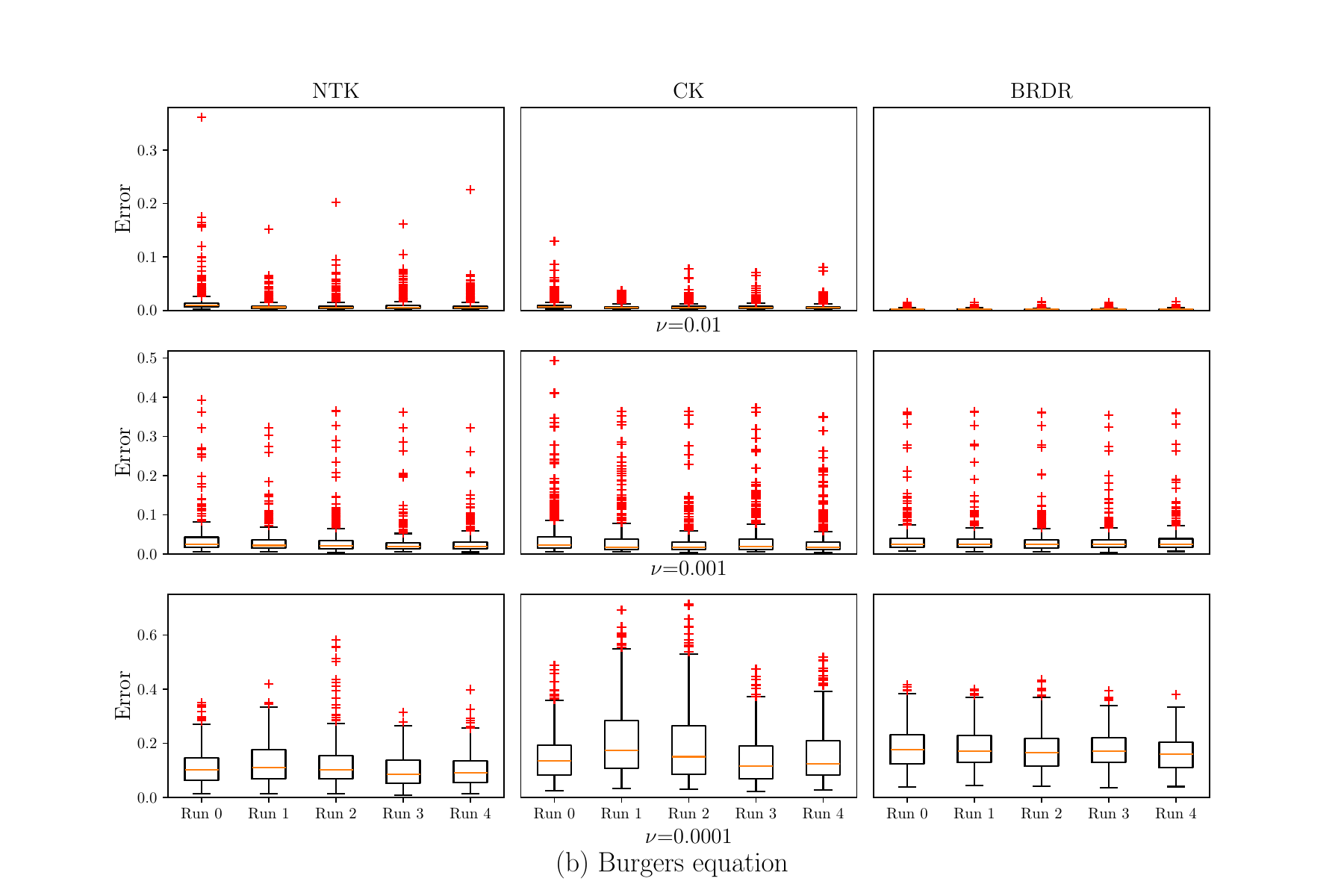}
		}
		\caption{PIDeepONets for the 1D wave equation and 1D Burgers equation: box-plots of prediction errors over different initial conditions from various runs. The points denoted as red crosses are outliers, positioned beyond the whiskers which extend to 1.5 times the inter-quartile range from the quartiles.
		}
		\label{Fig_ErrorBox_BurgersOperator}	
	\end{figure}

	The prediction errors are presented in Table \ref{tab_err_PIDeepONet}. The errors for all adaptive methods (NTK, CK, and BRDR) are smaller than those for fixed weights, underscoring the advantages of adaptive weighting. Compared to NTK and CK training, BRDR achieves significantly smaller prediction errors for $\nu = 0.01$. For $\nu = 0.001$, the prediction errors are comparable to those of NTK and CK training. For $\nu = 0.0001$, the prediction errors are larger than those of NTK and CK training. A possible reason for the differences  in prediction errors  especially at smaller viscosity is that different adaptive weighting methods emphasize different components of the loss function during training. When the training loss is large especially at smaller viscosity, the predictions tend to stay far from the ground truth. It is difficult to determine which aspect of the loss function to focus on to achieve smaller errors, as the effectiveness of each method can vary. Beyond prediction error, the significant advantages of BRDR training include much lower uncertainty, as discussed in Section \ref{sec_DeepONet_Wave}. 
	Similarly, to examine the distribution of prediction errors across different initial conditions,  we present box plots of the prediction errors from various runs in Fig. \ref{Fig_ErrorBox_BurgersOperator}(b). These plots also show that the BRDR method consistently achieves better uniformity of prediction errors compared to both NTK and CK methods, with only slight deviations. This increased uniformity is likewise attributed to the built-in uniformity of convergence rate in the BRDR method.
	Furthermore, as shown in Table \ref{tab_err_PIDeepONet}, BRDR training demonstrates considerable advantages in terms of computational time cost, similar to the results observed for the wave equation in Section \ref{sec_DeepONet_Wave}.

	\section{Conclusion}
	In conventional physics-informed machine learning, specifically in the area of physics-informed neural networks (PINNs) and physics-informed deep operator networks (PIDeepONets), the loss function is a linear combination of the squared residuals for the PDE, the BC, and the IC, each weighted with fixed coefficients. Training with fixed weights can sometimes result in significant discrepancies in the convergence rate of residuals at different training points. In this work, we introduce the concept of the \enquote{inverse residual decay rate} to describe the convergence rate of residuals. Based on this concept, we design an adaptive weighting method aimed at balancing the residual decay rate throughout the training process. In this method, the mean of all pointwise weights (positive) is constrained to be 1, ensuring that the weights remain bounded. Additionally, we use a scaling factor to keep the learning rate close to its maximum, thereby accelerating the training process.
	
	The performance of our proposed adaptive weighting method is compared with state-of-the-art adaptive weighting methods on benchmark problems for both PINNs and PIDeepONets. For PINNs, we compare the proposed adaptive weighting method with the recently proposed soft-attention weighting method and the residual-based attention weighting method. The test results show that the proposed method is characterized by high prediction accuracy and fast convergence rate, achieving not only a lower final prediction error but also a faster convergence rate. Moreover, we consistently use the same hyperparameters across different benchmarks, indicating an easy configuration for the proposed adaptive weighting method. For PIDeepONets, we compare the proposed adaptive weighting method with the recently proposed neural tangent kernel-based weighting method and the conjugate kernel-based weighting method. Except for the case of Burgers equation with the smallest viscosity tested, our method achieves a lower or comparable prediction error. The proposed method is particularly notable for its significantly lower uncertainty and much lower computational cost.

	Our adaptive weighting method is formulated to balance the convergence rate of residuals, leveraging the fact that residuals decay exponentially during the training process of neural networks. This approach can be extended to other deep learning frameworks, provided that the residuals also exhibit exponential decay. Furthermore, our tests on the Burgers equation with PIDeepONet indicate that incorporating adaptive sampling is essential for further enhancing the effectiveness of adaptive weighting during training.
	
	\section*{Acknowledgments}
	The work is  supported by the U.S. Department of Energy, Advanced Scientific Computing Research
	program, under the Scalable, Efficient and Accelerated Causal Reasoning Operators, Graphs and Spikes for Earth and
	Embedded Systems (SEA-CROGS) project (Project No. 80278).
	Pacific Northwest National Laboratory (PNNL) is a multi-program national laboratory operated for the U.S. Department of Energy (DOE) by Battelle Memorial Institute under Contract No. DE-AC05-76RL01830.

	\appendix
	\section{Network architectures}\label{sec_net}
	The modified fully-connected network (mFCN) introduced in \cite{wang2021understanding}, which has demonstrated to be more effective than the standard fully-connected neural network.
	A mFCN maps the input $\mathbf{x}$ to the output $\mathbf{y}$. Generally, a mFCN consists of an input layer, $L$ hidden layers and an output layer.  The $l$-th layer has $n_l$ neurons, where $l=0,1,..L,L+1$ denotes the input layer, first hidden layer,..., $L$-th hidden layer and the output layer, respectively. Note that the number of neurons of each hidden layer is the same, i.e., $n_1=n_2=...=n_L$. The forward propagation, i.e. the function $\mathbf{y}=f_{\pmb{\theta}}(\mathbf {x})$,  is defined as follows
	\begin{equation}
		\begin{aligned}
			\mathbf{U}&= \phi(\mathbf{W}^U\mathbf{x}+\mathbf{b}^U) &&\\
			\mathbf{V}&= \phi(\mathbf{W}^V\mathbf{x}+\mathbf{b}^V) &&\\
			\mathbf{H}^{1} &= \phi( \mathbf{W}^{1}\mathbf{x}+ \mathbf{b}^{1}) &&\\
			\mathbf{Z}^{l} &= \phi( \mathbf{W}^{l}\mathbf{H}^{l-1}+ \mathbf{b}^{l}), && 2\le l \le L \\
			\mathbf{H}^{l} &= (1-\mathbf{Z}^{l}) \odot \mathbf{U} + \mathbf{Z}^{l} \odot \mathbf{V}, && 2\le l \le L \\
			f_{\pmb{\theta}}(\mathbf{x})   &=\mathbf{W}^{L+1}\mathbf{H}^{L}+ \mathbf{b}^{L+1} &&\\
		\end{aligned}
		,
	\end{equation}
	where  $\phi(\bullet)$ is a pointwise activation and $\odot$ denotes pointwise multiplication. The training parameter in the network is $\pmb{\theta}=\{\mathbf{W}^U,\mathbf{W}^V, \mathbf{b}^U, \mathbf{b}^V, \mathbf{W}^{1:L+1}, \mathbf{b}^{1:L+1} \}$.

	The modified deep operator network (mDeepONet), inspired by the modified Fully Connected Network (mFCN) \cite{wang2021understanding}, is introduced in \cite{wang2022improved}. It has been shown to uniformly outperform the standard DeepONet architecture \cite{lu2021learning}. A DeepONet consists of two sub-networks: the trunk network and the branch network. The trunk network takes coordinates $\mathbf{x}$ as input, while the branch network takes a function (represented as $\mathbf{u}$) as input. The output of DeepONet is the inner product of the outputs of the trunk and branch networks. Considering the trunk and branch networks both have $L$ hidden layers, the forward propagation, i.e., the function $\mathbf{y}=G_{\pmb{\theta}}(\mathbf{u})(\mathbf{x})$, is defined as follows:
	\begin{equation}
		\begin{alignedat}{6}
			\mathbf{U} &= \phi(\mathbf{W}_{\mathbf{u}}\mathbf{u}+\mathbf{b}_{\mathbf{u}}) &\qquad& 
			\mathbf{V} &&= \phi(\mathbf{W}_{\mathbf{x}}\mathbf{x}+\mathbf{b}_{\mathbf{y}}) &&\\
			\mathbf{H}_{\mathbf{u}}^{1} &= \phi( \mathbf{W}_{\mathbf{u}}^{1}\mathbf{x}+ \mathbf{b}_{\mathbf{u}}^{1}) &&
			\mathbf{H}_{\mathbf{x}}^{1} &&= \phi( \mathbf{W}_{\mathbf{x}}^{1}\mathbf{x}+ \mathbf{b}_{\mathbf{x}}^{1}) \\
			\mathbf{Z}_{\mathbf{u}}^{l} &= \phi( \mathbf{W}_{\mathbf{u}}^{l}\mathbf{H}_{\mathbf{u}}^{l-1}+ \mathbf{b}_{\mathbf{u}}^{l}) &&
			\mathbf{Z}_{\mathbf{x}}^{l} &&= \phi( \mathbf{W}_{\mathbf{x}}^{l}\mathbf{H}_{\mathbf{x}}^{l-1}+ \mathbf{b}_{\mathbf{x}}^{l}) \qquad  & 2\le l \le L \\		
			\mathbf{H}_{\mathbf{u}}^{l} &= (1-\mathbf{Z}_{\mathbf{u}}^{l}) \odot \mathbf{U} + \mathbf{Z}_{\mathbf{u}}^{l} \odot \mathbf{V} &&
			\mathbf{H}_{\mathbf{x}}^{l} &&= (1-\mathbf{Z}_{\mathbf{x}}^{l}) \odot \mathbf{U} + \mathbf{Z}_{\mathbf{x}}^{l} \odot \mathbf{V} \qquad  &2\le l \le L \\		
			\mathbf{H}_{\mathbf{u}}^{L+1} &=\mathbf{W}_{\mathbf{u}}^{L+1}\mathbf{H}_{\mathbf{u}}^{L}+ \mathbf{b}_{\mathbf{u}}^{L+1} &&
			\mathbf{H}_{\mathbf{x}}^{L+1}&&=\mathbf{W}_{\mathbf{x}}^{L+1}\mathbf{H}_{\mathbf{x}}^{L}+ \mathbf{b}_{\mathbf{x}}^{L+1} &\\
			G_{\pmb{\theta}}(\mathbf {u})(\mathbf {x}) &= \mathbf{H}_{\mathbf{u}}^{L+1} \cdot \mathbf{H}_{\mathbf{x}}^{L+1} &&&& & 
		\end{alignedat}
		,
	\end{equation}
	where the training parameter is $\pmb{\theta}=\{\mathbf{W}_{\mathbf{u}}, \mathbf{b}_{\mathbf{u}}, \mathbf{W}{\mathbf{u}}^{1:L+1}, \mathbf{b}_{\mathbf{u}}^{1:L+1}, \mathbf{W}_{\mathbf{x}}, \mathbf{b}_{\mathbf{x}}, \mathbf{W}_{\mathbf{x}}^{1:L+1}, \mathbf{b}_{\mathbf{x}}^{1:L+1}\}$.

		\section{Ablation study}\label{sec_ablation}
\begin{figure}[htbp]
	\centering	
	\includegraphics[ width=15cm, trim=1cm 0cm 1cm 0cm]{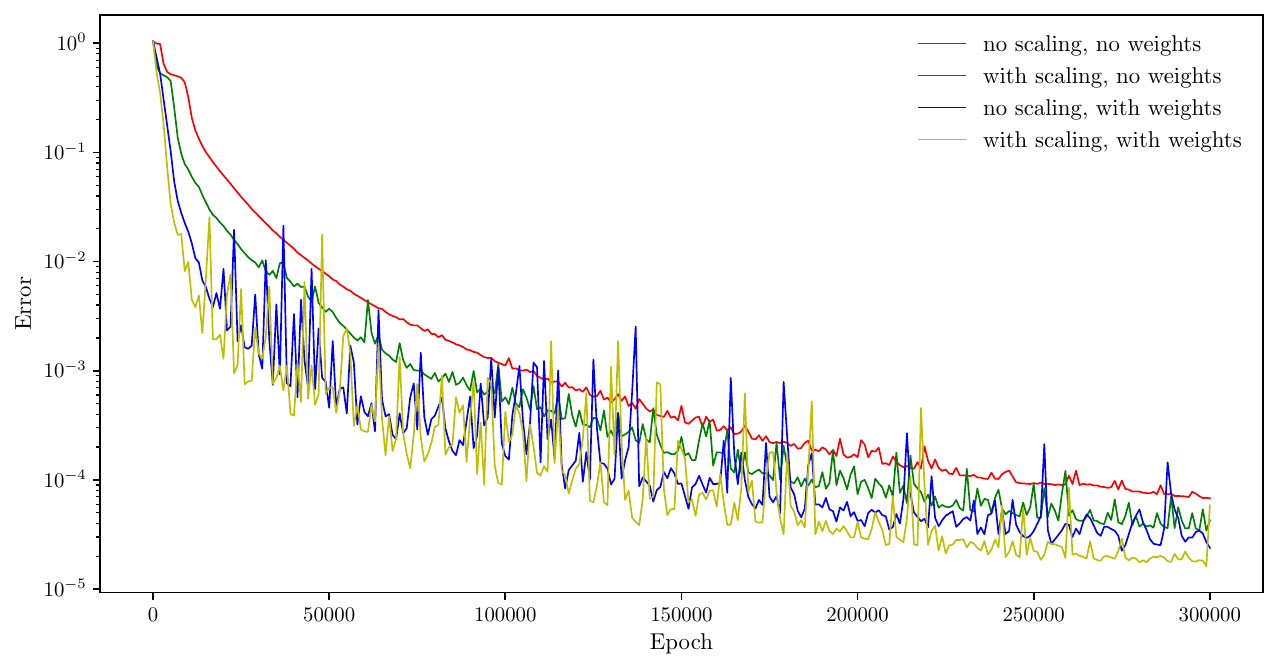}
	\caption{Error history of PINN predictions for Allen-Cahn equation. For brevity, the legend abbreviates \enquote{scaling factor} as \enquote{scaling} and \enquote{pointwise weights} as \enquote{weights}. Note that Fourier feature and the modified fully-connected network is included in the network architecture. }
	\label{Fig_errorHistory_AllenCahn_Ablation}	
    \end{figure}

		The proposed BRDR method comprises three distinct components: pointwise weights in Section \ref{sec_brdr}, scaling factor in Section \ref{sec_max_lr}, and mini-batch training in Section \ref{sec_mini_batch}, implemented in specific sections of our framework. Additionally, in our test cases, we integrate components from existing literature, including a modified fully-connected network (mFCN) and Fourier feature embedding. To evaluate the contributions of each component, we conduct an ablation study by selectively omitting one or more components during each trial on the Allen-Cahn equation described in Section \ref{sec_AllenCahn}. As mini-batch training is not employed for the Allen-Cahn equation, we  excluded it from the ablation study. For the ablation study, in the absence of mFCN, a standard FCN with an equivalent number of layers and neurons serves as the baseline for comparison. Including the Fourier feature automatically satisfies the periodic boundary conditions, thus eliminating the need for its component loss in the total loss function. Conversely, when the Fourier feature is excluded, we enforce the periodic boundary conditions by incorporating the component loss into the total loss function. For this purpose, we uniformly sample \(N_B = 200\) boundary points across \((x,t) \in \{0,1\} \times [0,1]\) to calculate the periodic boundary condition loss, subsequently reformulating the total loss function as follows:
		\begin{equation}
			\mathcal{L}(\pmb{\theta}; \mathbf{w}, s, \pmb{\alpha}) = s\left( \frac{\alpha_R}{N_R}\sum_{i=1}^{N_R}w_R^i\mathcal{R}^2(\mathbf{x}_R^i)
			+\frac{\alpha_R}{N_B}\sum_{i=1}^{N_B}w_B^i\mathcal{B}^2(\mathbf{x}_B^i)
			+\frac{\alpha_I}{N_I}\sum_{i=1}^{N_I}w_I^i\mathcal{I}^2(\mathbf{x}_I^i)
			\right)
		\end{equation}
		where the weight constants $\pmb{\alpha}=1$ is employed.
		To ensure a fair comparison, the best-performing seed of the BRDR method, as shown in Table \ref{tab_err_PINN}, is selected for all test cases in the ablation study. The prediction errors for all test cases are listed in Table \ref{table_ablation_AllenCahn}. Compared to the baseline standard FCN, which lacks advanced components, the addition of mFCN reduces the error by a factor of 3 to 50, as evidenced by the comparison of case pairs 1–9, 2–10, 3–11, 4–12, and 8-16. Similarly, the incorporation of Fourier feature embedding decreases the error by a factor of 3 to 100, as shown by the comparison of case pairs 4-8, 9–13, 10–14, 11–15, and 12–16. The application of the BRDR method reduces the error by a factor of 5 to 15, as demonstrated by the comparison of case pairs 1–4, 9–12, and 13–16. Implementing only the scaling factor reduces the error by a factor of 1 to 3, as indicated by case pairs 1–2, 3–4, 9–10, 11–12, 13–14, and 15–16. Likewise, implementing only pointwise weights reduces the error by a factor of 2 to 10, as shown by case pairs 1–3, 2–4, 9–11, 10–12, and 13–15, 14–16. Figure \ref{Fig_errorHistory_AllenCahn_Ablation} illustrates the history of testing error when mFCN and Fourier feature embedding are active, demonstrating that the inclusion of BRDR components also speeds up convergence.
		
		Besides, in our experiments (cases 5, 6, and 7), we observed that when the Fourier feature embedding is used without mFCN, the network fails to converge. We suspect this issue arises from the improper application of the Fourier feature embedding. Specifically, the initial condition, $ u(x,0) = x^2 \cos(\pi x)$ in Eq. \eqref{eq_AC}, is not strictly periodic on the interval $[-1, 1]$ because the first-order derivatives at $ x = -1$ and $ x = 1 $ do not match. As a result, the network get trapped into a solution that is significantly different from the exact solution, with the training loss remaining high. In contrast, the inclusion of  mFCN can successfully overcome this issue. The BRDR method—with its combination of scaling factor and pointwise weights—also helps alleviate the issue. While our empirical results demonstrate that both mFCN and the BRDR method improve convergence, the  mechanisms behind their effectiveness remain unclear and warrant further investigation.
		
		Overall, integrating both mFCN and Fourier feature embedding is essential for significantly reducing the prediction error, and further inclusion of the BRDR method can push the error down to a minimal level.

		Regarding training time, incorporating mFCN results in a significant increase in training time. In contrast, integrating Fourier feature embedding reduces training time by approximately 10\% by eliminating the need to calculate the boundary loss. Meanwhile, the addition of BRDR components incurs an increase of less than 10\% in training time.

	\begin{table}[tbp]
		\newcommand{\tabincell}[2]{\begin{tabular}{@{}#1@{}}#2\end{tabular}}
		\centering
		\caption{Relative $L_2$ prediction error and relative training time cost for each case of the 
			ablation study for the 1D Allen–Cahn Equation. The symbol 
			\ding{51} indicates that the corresponding component is included in the case, whereas \ding{55} denotes that the component is excluded.}
		\begin{tabular}{ccccccc}
			\toprule
			\multirow{2}{*}{} & \multicolumn{4}{c}{Components}                                                                                                                   & \multirow{2}{*}{Error} & \multirow{2}{*}{Time}\\
			\cmidrule(lr{8pt}){2-5}
			& mFCN & Fourier & \begin{tabular}[c]{@{}c@{}}Scaling\\ factor\end{tabular} & \begin{tabular}[c]{@{}c@{}}Pointwise\\ weights\end{tabular} &               &         \\ \midrule
			Case 1    & \ding{55}   & \ding{55}      & \ding{55} & \ding{55} & 1.35e-2          & 100\%       \\ 
			Case 2    & \ding{55}   & \ding{55}      & \ding{51} & \ding{55} & 1.38e-2          & 105\%       \\ 
			Case 3    & \ding{55}   & \ding{55}      & \ding{55} & \ding{51} & 6.00e-3          & 105\%       \\ 
			Case 4    & \ding{55}   & \ding{55}      & \ding{51} & \ding{51} & 2.14e-3          & 106\%       \\ 
			Case 5    & \ding{55}   & \ding{51}      & \ding{55} & \ding{55} & 9.80e-1          & 89\%       \\ 
			Case 6    & \ding{55}   & \ding{51}      & \ding{51} & \ding{55} & 9.94e-1          & 97\%       \\ 
			Case 7    & \ding{55}   & \ding{51}      & \ding{55} & \ding{51} & 9.97e-1          & 95\%       \\ 
			Case 8    & \ding{55}   & \ding{51}      & \ding{51} & \ding{51} & 8.87e-4          & 102\%       \\ 
			Case 9    & \ding{51}   & \ding{55}      & \ding{55} & \ding{55} & 3.48e-3          & 176\%       \\ 
			Case 10   & \ding{51}   & \ding{55}      & \ding{51} & \ding{55} & 3.19e-3          & 182\%       \\ 
			Case 11   & \ding{51}   & \ding{55}      & \ding{55} & \ding{51} & 2.36e-4          & 179\%       \\ 
			Case 12   & \ding{51}   & \ding{55}      & \ding{51} & \ding{51} & 1.92e-4          & 185\%       \\ 
			Case 13   & \ding{51}   & \ding{51}      & \ding{55} & \ding{55} & 6.72e-5          & 165\%       \\ 
			Case 14   & \ding{51}   & \ding{51}      & \ding{51} & \ding{55} & 3.39e-5          & 171\%       \\ 
			Case 15   & \ding{51}   & \ding{51}      & \ding{55} & \ding{51} & 2.38e-5          & 165\%       \\ 
			Case 16   & \ding{51}   & \ding{51}      & \ding{51} & \ding{51} & \bf{1.60e-5}     & 171\%       \\ 				
			\bottomrule             
		\end{tabular}		
		\label{table_ablation_AllenCahn}
	\end{table}

	\bibliographystyle{elsarticle-num}
	\bibliography{bibliography}
	
\end{document}